\documentclass[runningheads]{llncs}
\newcommand{\myparagraph}[1]{{\noindent \bf #1}}
\usepackage{graphicx}
\usepackage{comment}
\usepackage{amsfonts,amsmath,amssymb} % define this before the line numbering.
\usepackage{color}
\usepackage{mathrsfs}
\usepackage{subfigure}
\usepackage{multirow}
\usepackage{caption2}
\usepackage[misc]{ifsym}

\usepackage{xcolor}
\usepackage[pagebackref=false, breaklinks=true, letterpaper=true, colorlinks,citecolor=citecolor, linkcolor=linkcolor, bookmarks=false]{hyperref}
\definecolor{citecolor}{HTML}{0071BC}
\definecolor{linkcolor}{HTML}{ED1C24}

\usepackage[capitalize]{cleveref}
\crefname{section}{Sec.}{Secs.}
\Crefname{section}{Section}{Sections}
\Crefname{table}{Table}{Tables}
\crefname{table}{Tab.}{Tabs.}

\newcommand{\etal}{\textit{et al}.~}
\newcommand{\ie}{\textit{i}.\textit{e}.~}
\newcommand{\ieno}{\textit{i}.\textit{e}.}

\newcommand{\egno}{\textit{e}.\textit{g}.}
\usepackage{multirow}
\usepackage{ulem}

\usepackage[accsupp]{axessibility}  % Improves PDF readability for those with disabilities.

\begin{document}
% \linenumbers
\pagestyle{headings}
\mainmatter
\def\ECCVSubNumber{1954}  % Insert your submission number here

\title{
Image Coding for Machines with Omnipotent Feature Learning
} % Replace with your title

% % INITIAL SUBMISSION 
% %\begin{comment}
% \titlerunning{ECCV-22 submission ID \ECCVSubNumber} 
% \authorrunning{ECCV-22 submission ID \ECCVSubNumber} 
% \author{Anonymous ECCV submission}
% \institute{Paper ID \ECCVSubNumber}
% %\end{comment}
% %******************

% \begin{comment}
\titlerunning{Image Coding for Machines with Omnipotent Feature Learning}
% \author{Ruoyu Feng\inst{1}\orcidID{0000-1111-2222-3333} \and
\author{{Ruoyu Feng\inst{1*}}\quad
Xin Jin\inst{2*}\quad
Zongyu Guo\inst{1}\quad
Runsen Feng\inst{1}\quad
Yixin Gao\inst{1}\quad
\\Tianyu He\inst{3}\quad
Zhizheng Zhang\inst{3}\quad
Simeng Sun\inst{1}\quad
Zhibo Chen\inst{1,\dag}
}
\authorrunning{Ruoyu Feng et al.}
\institute{University of Science and Technology of China
\and
Eastern Institute of Advanced Study
\and
Microsoft Research Asia, Beijing, China\\
\email{ustcfry@mail.ustc.edu.cn\quad jinxin@eias.ac.cn\quad\\
chenzhibo@ustc.edu.cn}
}
% \end{comment}
%******************
\maketitle
\let\thefootnote\relax\footnotetext{* First two authors contributed equally.\\
\dag~Corresponding author.}

\begin{abstract}

Image Coding for Machines (ICM) aims to compress images for AI tasks analysis rather than meeting human perception. Learning a kind of feature that is both general (for AI tasks) and compact (for compression) is pivotal for its success. In this paper, we attempt to develop an ICM framework by learning universal features while also considering compression. We name such features as omnipotent features and the corresponding framework as Omni-ICM. Considering self-supervised learning (SSL) improves feature generalization, we integrate it with the compression task into the Omni-ICM framework to learn omnipotent features. However, it is non-trivial to coordinate semantics modeling in SSL and redundancy removing in compression, so we design a novel information filtering (IF) module between them by co-optimization of instance distinguishment and entropy minimization to adaptively drop information that is weakly related to AI tasks (\egno, some texture redundancy). Different from previous task-specific solutions, Omni-ICM could directly support AI tasks analysis based on the learned omnipotent features without joint training or extra transformation. Albeit simple and intuitive, Omni-ICM significantly outperforms existing traditional and learning-based codecs on multiple fundamental vision tasks.

\end{abstract}

\section{Introduction}
\vspace{-1mm}

% Video coding, which targets to compress and reconstruct the whole frame, and feature compression, which only
% preserves and transmits the most critical information, stand at
% two ends of the scale

% In the big data era, massive images and videos are generated and transmitted among people everyday\fry{(images and videos have become an indispensable part of people's production and life.)}.
In the big data era, massive images and videos have become an indispensable part of people's production and life.
As an important industrial technology, lossy image compression aims to save storage resources and transmission bandwidth by preserving the most critical information.
In the past decades, the traditional image and video coding standards such as JPEG~\cite{wallace1992jpeg}, JPEG2000~\cite{rabbani2002overview}, AVC/H.264~\cite{wiegand2003overview}, HEVC/H.265~\cite{sullivan2012overview}, VVC/H.266~\cite{bross2021overview} have significantly improved the coding efficiency. 
Recently, with the fast development of deep neural networks, learning-based image compression codecs~\cite{balle2017end,balle2018variational,minnen2018joint,cheng2020learned,johnston2018improved,li2018learning,li2020learning,mentzer2018conditional,mentzer2020high,chen2019learning,wu2021learned} have achieved a great success. They have potentials to become the next-generation image compression standards due to the high performance and applicability compared to traditional hand-craft codecs.
Meanwhile, deep neural networks has demonstrated their potential in various computer vision tasks, \egno, object detection~\cite{ren2015faster,redmon2016you,redmon2017yolo9000,lin2017feature}, instance segmentation~\cite{he2017mask,liu2018path,bolya2019yolact}, semantic segmentation~\cite{long2015fully,badrinarayanan2017segnet,chen2017deeplab,chen2018encoder}, pose estimation~\cite{he2017mask,newell2016stacked}. We can anticipate that more and more data transmitting on the Internet would be consumed by machines for intelligent analysis tasks. 

\vspace{-6mm}
  
\begin{figure}
  \centerline{\includegraphics[width=1.0\linewidth]{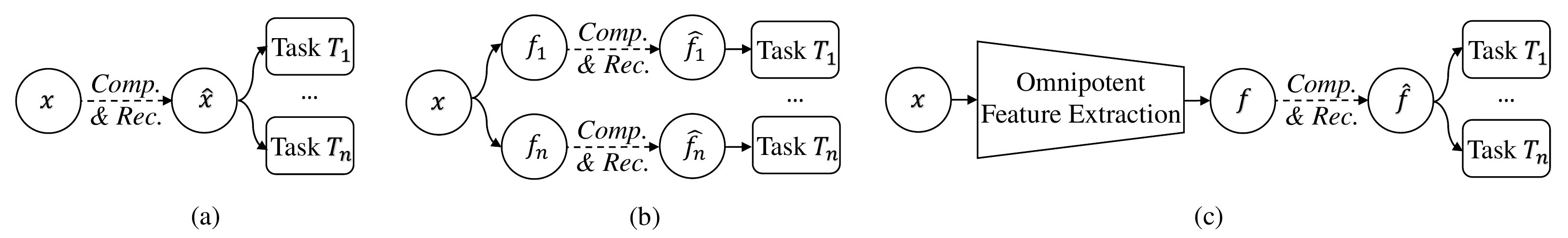}}
  \vspace{-3mm}
    \caption{Comparison of three branches for image coding for machines (ICM). They are different from each other w.r.t the object to be compressed and the characteristics of task-specific or not. {(a)}: Codecs in this branch support downstream tasks by inputting the decompressed images.
     {(b)}: One-to-one features-based ICM solution, the decompressed features of corresponding tasks are input to the task models. 
     {(c)}: With the proposed \textbf{omnipotent feature} $f$ extracted and compressed first, all the downstream tasks could complete the inference based on the decompressed feature $\hat{f}$.
    %  {(c)}: The proposed \textbf{omnipotent feature} $f$ is extracted and compressed first, and then all the downstream tasks could complete the inference based on the universal decompressed $\hat{f}$.
    %  which solution significantly improves generalization and flexibility.
     }
\label{fig:motivation}
\vspace{-4mm}
\end{figure}

\vspace{-2mm}

However, all the image compression methods mentioned above aim at saving transmitting costs while improving the reconstruction quality for human perception. 
% When facing big data and AI task analytics, existing image coding methods (even for the deep learning-based) are still questionable, regarding whether such big data can be efficiently handled by visual signal level compression.
When facing AI tasks analysis, existing image coding methods (even for the deep learning-based) are still questionable, regarding whether it can encode images efficiently, especially in application scenarios for big data.
To facilitate the performance and efficiency in terms of high-level machine vision tasks that act on lossy compressed images, lots of research efforts have been
dedicated to a new problem of image coding for machines (ICM)~\cite{duan2020video,le2021image}, which aims to compress the source image for supporting the intelligent analysis tasks. The discrepancy between human-perception oriented metric (\egno, mean square error (MSE), multi-scale structured similarity (MS-SSIM)) and AI task metric (\egno, classification accuracy) makes ICM particularly different from the existing compression schemes.

For ICM, there mainly exist solutions of two branches. Fig. \ref{fig:motivation}(a) shows the first branch that the compressed image is sent into the downstream task model for intelligent analytics. Codecs in this branch are typically designed based on a heuristic RoI (Region of Interest) bit allocation strategy~\cite{song2021variable,choi2018high,cai2021novel,huang2021visual} or joint optimization for image reconstruction with a task-specific constraint in an end-to-end manner~\cite{le2021image}. 
This branch has two weaknesses that the image reconstruction brings more computational burden 
because images have to be reconstructed for subsequent intelligent analysis and there exists a new trade-off between texture fidelity and semantics integrity.
The second branch is a one-to-one feature-based ICM framework~\cite{chen2019lossy,chen2019toward,singh2020end,bajic2021collaborative}. 
% As shown in Figure~\ref{fig:motivation} (b), works of this branch tend to extract features from images, and then compress them for transmission efficiency.
As shown in Fig.~\ref{fig:motivation}(b), works of this branch tend to compress the features extracted from images for transmission efficiency.
Depending on the reconstructed features, the downstream tasks could directly complete the corresponding intelligent analysis.
But, such a scheme that one compressed feature can only be used to support one specific AI task lacks generalization and flexibility, thus is difficult to be applied to practical applications.

To solve the problems mentioned above, and motivated by the urgent requirements for a generalized ICM solution, in this paper, we go beyond previous pipelines and introduce a unified framework for ICM by exploring the ``common knowledge'' of different AI tasks. 
More precisely, a novel ICM framework, termed Omni-ICM, is designed based on learning omnipotent features for machines, as shown in Fig.~\ref{fig:motivation}(c). 
The omnipotent features are expected to be general for different intelligent tasks and compact enough that only contain the semantics relevant information. 
They can be regarded as new representations ``seen'' by machines. To achieve the omnipotent feature learning, we borrow ideas from the popular contrastive learning that has been proved could learn general and transferable visual representations~\cite{he2020momentum,chen2020simple,caron2020unsupervised,grill2020bootstrap,chen2021exploring}, and integrate it into the image coding pipeline. 
However, directly compressing the features learned by contrastive objective has no obvious advantages than compressing the original images directly~\cite{chen2020data,chen2019lossy,chen2018intermediate}, that's because these features typically keep lots of irrelevant redundant information with no explicit constraint on information entropy.

To tackle this issue, we further design an Information Filtering (IF) module to smartly discard the redundant information for analytics before compression, so as to encourage learned representations to be sparse and compact. 
Basically, the IF module comprises an encoder, a decoder, and an entropy estimation model, and is optimized with contrastive loss and entropy minimization constraint. In this way, IF module learns to preserve semantic-wise information and filter out redundant ones, acting as a bridge to connect contrastive training and compression. 
After that, with a learning-based feature compressor, the learned omnipotent features are compressed and reconstructed in the feature latent space, enabling it to be directly input to downstream task models without pixel-level reconstruction.
% which makes it possible to be robust and generate less semantic distortion during the compression process. 
Moreover, compressing such omnipotent features makes it more applicable to the codec standardization, which could support for a wide range of downstream AI tasks, even for the unknown ones. Such generalization ability and flexibility are the key points of our Omni-ICM framework, which are often neglected by the existing ICM solutions. 

Extensive experiments show that Omni-ICM outperforms the state-of-the-art image compression methods by significant margins w.r.t the bitstream saving and task performance, on multiple intelligent tasks, including object detection, instance/semantic/panoptic segmentation, and pose estimation. 

\vspace{-1mm}
\section{Related Work}
\vspace{-1mm}
\subsection{Image Compression}
\vspace{-1mm}
\myparagraph{Traditional Codec.} 
% From the 1970s, the hybrid video coding architecture~\cite{roese1975combined} is proposed to lead the mainstream direction and occupy the major industry proportion during the next few decades. 
Traditional hand-craft image codecs typically consist of intra prediction, discrete cosine transformation or wavelet transformation, quantization, and entropy coder. 
The popular image coding standards have kept evolving, \egno, JPEG~\cite{wallace1992jpeg}, JPEG2000~\cite{rabbani2002overview}, AVC~\cite{wiegand2003overview}, HEVC/H.265~\cite{sullivan2012overview}, VVC/H.266~\cite{bross2021overview}. However, these codecs cannot be optimized in an end-to-end manner, thus lack of flexibility and scalability to support different objectives, such as MS-SSIM and classification accuracy.

% Each module is designed with multiple modes and rate-distortion optimization would be conducted to determine the best mode. 

% Generally, this branch aims to learn the representative and powerful features via a hierarchical network and is jointly optimized with the reconstruction task for high efficient coding.

\myparagraph{Learning-based Codec.} The great success of deep learning techniques significantly promotes the
development of end-to-end learning-based codec. Toderici \textit{et al.}~\cite{toderici2015variable} apply a recurrent neural network (RNN) to end-to-end image compression, achieving a comparable performance with JPEG.
Ball{\'e} \textit{et al.}~\cite{balle2017end} further propose 
an end-to-end framework based on nonlinear transformation, generalized divisive normalization (GDN), noise-relaxed quantization, and their method outperforms JPEG 2000. Then a variational model with hyperprior is introduced to parameterize latent distribution with a zero-mean Gaussian distribution in~\cite{balle2018variational}. 
% Minnen \textit{et al.}~\cite{minnen2018joint} design a context model combined with a mean and scale hyperprior to autoregressively predict the latent representations from the decoded content,  which surpasses BPG in terms of rate-distortion performance.
% Cheng \textit{et al.}~\cite{cheng2020learned} replace the original single Gaussian model with a mixture one to get more accurate entropy estimation. 
Some recent works have improved image compression from the aspects of entropy coding~\cite{minnen2018joint,minnen2020channel,cheng2020learned,guo2021causal,kim2021joint} and quantization~\cite{guo2021soft,yang2020improving}.
% Kim \textit{et al.}~\cite{kim2021joint} exploits both local and global information in a content-dependent manner for entropy coding.
% Guo \textit{et al.}~\cite{guo2021causal}
% More recent works have improved image compression from the aspects of entropy coding \
However, the optimization objectives of these learning-based compression methods are pixel-level metrics that designed for visual fidelity, \egno, MSE, MS-SSIM. 
The discrepancy between pixel-level distortion and semantic-level distortion leads to the failure of above methods when tackling ICM tasks.
But, they provide basic techniques to develop effective ICM solutions to handle this new problem.

% In our study, during the omnipotent feature learning stage, we apply entropy estimation in the IF module by a factorized entropy model with additive noise\cite{balle2017end}.
% After that, a variational auto-encoder based compressor is trained to compress the omnipotent feature.

% For the traditional codecs, ROI (Region of Interest) based bit allocation\cite{cai2021novel, huang2021visual} is a heuristic direction by the intuition that foreground matters more than background for intelligent analytics

% For learning-based codecs that optimized in an end-to-end manner, 

% where the high-level semantiinformation is carefully designed such as body key points or face edges, serving for machine tasks.

\myparagraph{Image Coding for Machine.} ICM~\cite{duan2020video,he2019beyond} aims to compress and transmit the source image for machines to support intelligent tasks such as object detection, semantic segmentation.
% (such as object detection, semantic segmentation.)
Based on the heuristic prior knowledge of foreground matters more for intelligent analysis, \cite{cai2021novel,huang2021visual,li2021task} merge the ROI (Region of Interest) based bit allocation strategy into the traditional codec for intelligent analytics. 
% And Li \etal\cite{li2021task} explore the idea of task-driven bit-allocation based on reinforcement learning.
For learning-based codecs, Le \etal~\cite{le2021image} propose an image compression system that jointly optimizes models for object detection and reconstruction. 
Codevilla \etal~\cite{codevilla2021learned} also optimize both the intelligent task and the reconstruction task, and the difference is that the optimization of the intelligent task directly takes the latent variable features as input.
However, the trade-off between semantic fidelity and pixel fidelity limits their respective performance.
Thus, \cite{hu2020towards,xia2020emerging} introduce scalable coding ideas to coordinate the compression for high-level information and pixel-wise texture.
% by manually preserving key semantics for machine tasks.} 
Singh \etal~\cite{singh2020end} explore to compress features instead of images for intelligent tasks by optimizing the task objective along with rate loss.
Nevertheless, such schemes can only support a few tasks and are not general enough. 
Similarly, the recent work of SSIC~\cite{sun2020semantic} structures the bitstream according to the object category and thus achieves a task-aware decompression for downstream analytics. 
% Dubois \etal \cite{dubois2021lossy} trains a specific compressor for image classification for rate savings while without performance decrease. 
Differently, in this paper, we aim to design a unified framework for image coding for machines by learning a kind of general and compact features and directly support a wide range of intelligent tasks.

\vspace{-1mm}
\subsection{Self-supervised Representation Learning (SSL)}
\vspace{-1mm}
Self-supervised learning~\cite{jing2020self,zhang2016colorful,pathak2016context,noroozi2016unsupervised}
%as a subset of unsupervised learning methods, 
is proposed to learn general representations for downstream tasks by solving various pretext tasks on large-scale unlabeled datasets. 
% Contrastive learning is one of them and its pretext task is minimizing feature distances from the same group and maximizing feature distances from different groups with triplet loss or contrastive loss.
Contrastive learning is one of them and its pretext task is minimizing feature distances from the same group and maximizing feature distances from different groups with contrastive loss.
% Representatively,
Recently, the siamese network based contrastive learning methods~\cite{he2020momentum,chen2020simple,caron2020unsupervised,grill2020bootstrap,chen2021exploring} have drawn lots of attention. 
% Among them, MOCO~\cite{he2020momentum} is the first work that outperforms the supervised ImageNet pre-training on several downstream tasks, \egno, object detection, semantic segmentation, which shows its strong ability for general representations learning.
Among them, MOCO~\cite{he2020momentum} is the first work that outperforms the supervised ImageNet pre-training on several downstream tasks, which shows its strong ability for general representations learning.
More specifically, MOCO designs a dynamic queue to store negative samples features and uses a momentum update mechanism to optimize the model progressively. 
Inspired by that, we propose to employ SSL to learn omnipotent features for compression, so that further support heterogeneous intelligent tasks for ICM.

\vspace{-1mm}
\section{ICM with Omnipotent Feature Learning}
\vspace{-1mm}
\subsection{Overview of Omni-ICM Pipeline}
\vspace{-1mm}

\begin{figure}
  \centerline{\includegraphics[width=1.0\linewidth]{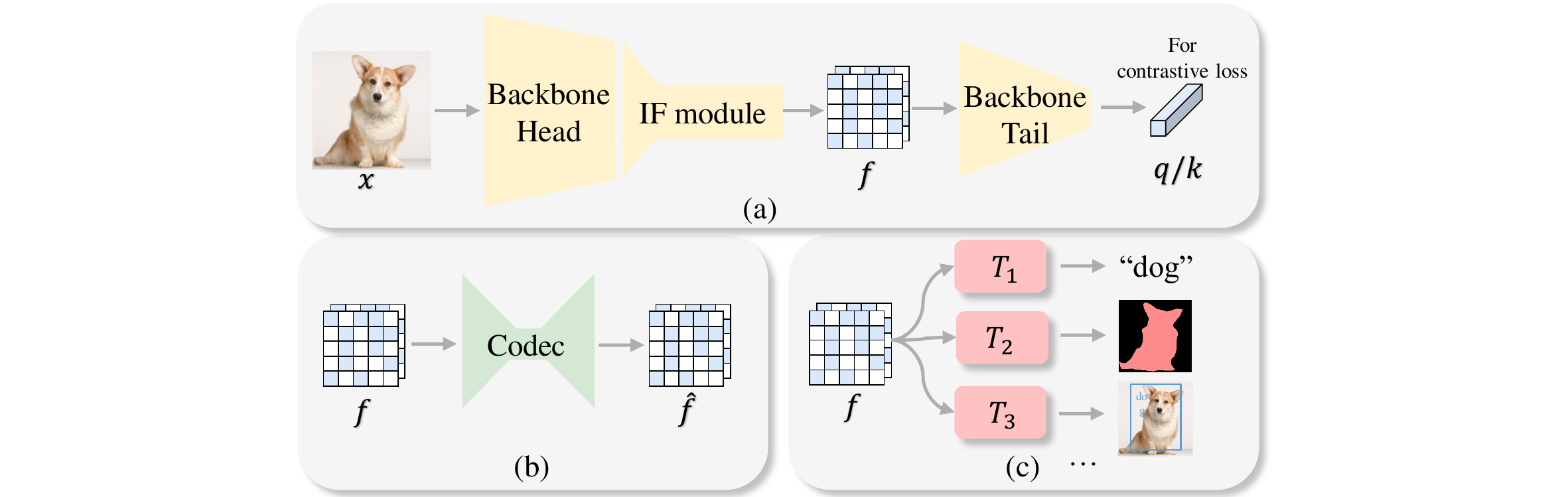}}
  \vspace{-3mm}
    \caption{Three stages in our Omni-ICM framework. (a) Omnipotent feature learning. We optimize the whole network with the contrastive loss and entropy constraint by the IF module. (b) Omnipotent feature compression. A feature compressor is trained for omnipotent feature compression, with all parameters fixed except the codec. (c) Omnipotent feature deployment. Our Omni-ICM can easily support different downstream tasks by fine-tuning the backbone tail with omnipotent features as input. 
    % In practical, omnipotent features after reconstruction is sent into the well-trained task model for inference.
    }
\label{fig:pipeline}
\vspace{-5mm}
\end{figure}

We propose a new concept of omnipotent feature learning for image coding for machines, and correspondingly design a unified framework (Omni-ICM) based on it. 
As shown in Fig. \ref{fig:pipeline}, the whole framework of Omni-ICM consists of three stages: (a) omnipotent feature learning, (b) omnipotent feature compression, and (c) omnipotent feature deployment.

\vspace{-5mm}

For the first stage, we employ a contrastive learning pipeline while also giving consideration to compression efficiency, enabling the learned features to be both semantically preserved and compact. 
More specifically, to coordinate the preserving of the semantics and the discarding of the semantic-irrelevant redundancy, we design an additional Information Filtering (IF) module and optimize the whole network with an instance-contrastive loss under entropy constraint.

After that, the obtained omnipotent features, which are compact and general, are ``seen'' by machines as an alternative for original images. To compress and transmit the omnipotent features, we additionally train a feature codec. Finally, the downstream tasks supporting are achieved by fine-tuning the backbone tail. Note that, the backbone head and the proposed IF module are fixed in this stage. We describe each stage in detail in the following subsections.

\begin{figure}
  \centerline{\includegraphics[width=1.0\linewidth]{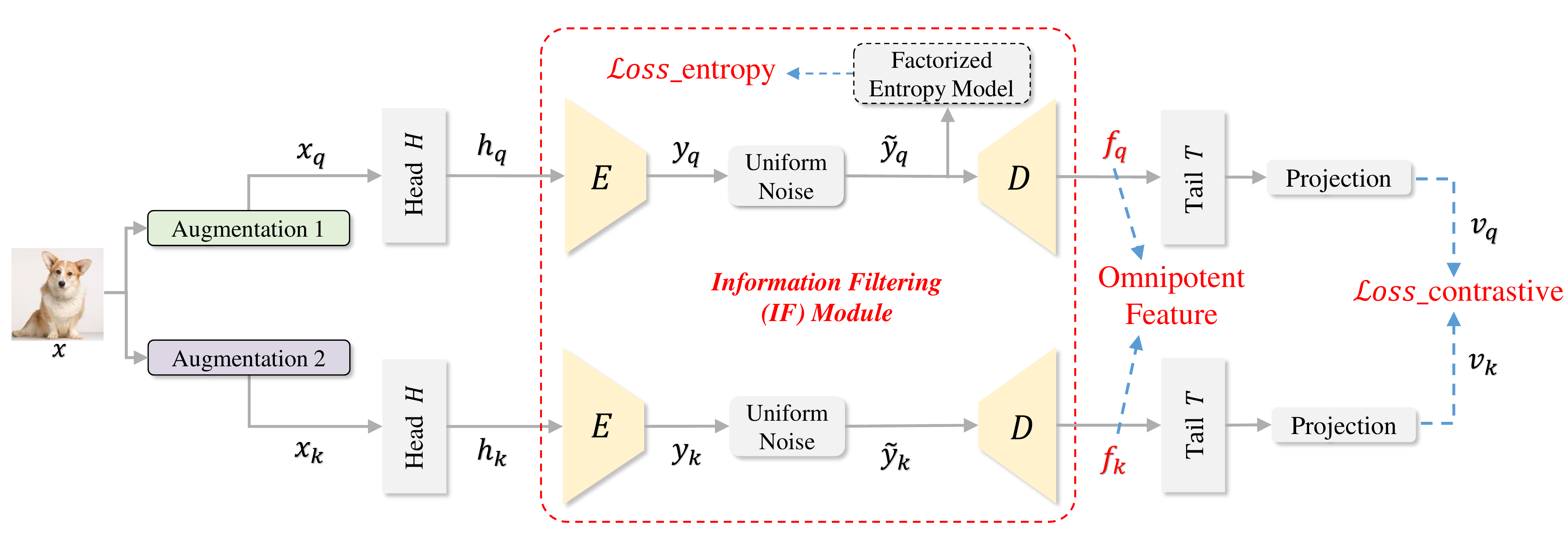}}
  \vspace{-4mm}
    \caption{Architecture of omnipotent feature learning. We use a pair of query and key for simpler illustration. 
    By maximizing the similarity of different views of an image under entropy constraint, the network learns to discard semantic-redundant information and keep critical ones. After training, $f$ is the omnipotent feature we need.}
\label{fig:feature_learning}
\vspace{-5mm}
\end{figure}

\vspace{-1mm}

\subsection{Stage 1: Omnipotent Feature Learning}
\label{section: omnipotent feature learning}
\vspace{-1mm}
\myparagraph{Basic Network Architecture.}
Considering that the learned omnipotent features will be taken for a wide range of AI tasks analytics, \egno, object detection~\cite{lin2017feature}, semantic segmentation~\cite{zhao2017pyramid}, we extract the omnipotent feature $f$ with a 4$\times$ down-sampling factor to promise the integrity of content structure and object spatial layout.
Specifically, as shown in Fig. \ref{fig:feature_learning}, a commonly used backbone (such as ResNet-50) is split into two parts, namely backbone head and backbone tail, {dotted} as $H$ and $T$.
In a ResNet-50, the backbone head comprises the stem layer and layer1, and the backbone tail comprises layer2$\sim$layer4.

\myparagraph{Data Augmentation and Feature Extraction in Backbone Head.} As illustrated in Fig. \ref{fig:feature_learning}, at the omnipotent feature learning stage, two views of an image $x_{q}$ and $x_{k}$ are first generated by different augmentations. For clarity, we describe the query generation process for $x_{q}$ at first. $x_{q}$ is fed into the backbone head $H$, obtaining an $4\times$ down-sampling feature with a size of $\frac{H_q}{4}\times\frac{W_q}{4}{\times}C$, where $H_q$, $W_q$ are the height and width of $x_{q}$, $C$ means the channel numbers:

\vspace{-3mm}
\begin{equation}
\small
    h_{q}=H(x_{q}).
\end{equation}
\vspace{-4mm}

\myparagraph{Information Filtering (IF) Module.} Importantly, the representation directly generated by the backbone head is not suitable for ICM, because it still contains lots of semantic-irrelevant information (see the third column of  Fig.~\ref{fig:qualitative_kodak23}).
Thus, we design an additional information filtering (IF) module between the backbone head and tail, to simultaneously achieve the preservation of semantic information and the dropout of irrelevant information. The IF module consists of an encoder, a factorized entropy model, and a decoder denoted as $E$, $F$, $D$. To drive the IF module to learn to filter out the redundant information, an entropy constraint is enforced on it.

% Details for training the whole framework are described as below.

Formally, $h_q$ is first fed into the encoder $E$ of IF module with $8\times$ down-sampling, obtaining a latent variable ${y}_{q}$ with the size of $\frac{H_q}{32}\times\frac{W_q}{32}{\times}C_y$, $C_y$ represents the channel numbers of $y_q$:

\vspace{-3mm}
\begin{equation}
    \small
    y_q=E(h_{q}).
\end{equation}
\vspace{-4mm}

Then, a factorized entropy model $F$ estimates the entropy of ${y}_{q}$ through adding an additive uniform noise\cite{balle2017end} on it to get the derivative $\tilde{y}_{q}$, formulated as:

\vspace{-3mm}
\begin{equation}
\begin{aligned}
&p_{\tilde{y}_{q}|{\phi_o}}(\tilde{y}_{q}|{\phi_o}) = \prod_{i} (p_{y_q|{\phi_o}}({\phi_o})\ast \mathcal{U}(-\frac{1}{2}, \frac{1}{2}))(\tilde{y}_{q}),
\end{aligned}
\end{equation}
where $\boldsymbol{\phi_o}$ represents the parameters in $H$ and $E$. 
% And, the entropy loss is setup by:
And, the entropy loss is:
\begin{equation}
\small
\mathcal{L}_{e} = \mathop{\mathbb{E}}[-\log_{2}(p_{{\tilde{{y}}}_{q}|{\phi_o}}({\tilde{{y}}}_{q}|{\phi_o} ))].
\end{equation}

Finally, $\Tilde{{y}}_{q}$ is fed into the decoder $D$ of IF module, obtaining the feature $f_q$ with the same size as the input of IF module, \ie $\frac{H_q}{4}\times\frac{W_q}{4}{\times}C$.

% , thus can be easily input into the backbone tail without any further conversion.

\myparagraph{Backbone Tail and Projection Layer.} 
With the feature $f_q$ generated by the IF module, the backbone tail and a projection layer are employed to map the feature to the space where contrastive loss is applied.
Specifically, the projection layer is an MLP with one hidden layer.
This procedure can be formulated as: 
\begin{equation}
    \small
    q=W^{(2)}\sigma(W^{(1)}(T(D(\tilde{y}_q)))),
\end{equation}
where $\sigma$ is a ReLU non-linearity transformation, $W^{(1)}$ and $W^{(2)}$ are fully connected layers, $q\in\mathop{\mathbb{R}}^d$.

\myparagraph{Generation of Keys.} 
$x_{k}$ is obtained by the other augmentation from the same image. The key $x_{k}$ and the query $x_{q}$ together construct a positive pair. 
% Following the setting in MOCO\cite{he2020momentum}, $x_{k}$ is updated with a momentum updating manner. 
For simplicity, we use the same notation in Section \ref{section: omnipotent feature learning} here. This procedure can be formulated as:
\begin{equation}
\small
    y_k=E(H(x_k)),
\end{equation}
\vspace{-4mm}
\begin{equation}
    \small
    k_{+}=W^{(2)}\sigma(W^{(1)}(T(D(\tilde{y}_k)))),
\end{equation}
where $\tilde{y}_k$ comes from $y_k$ by adding the additive uniform noise, and $k_{+}\in\mathop{\mathbb{R}}^d$, denotes the positive sample. The negative samples come from different images, denoted as$\{k_-\}$, are provided by the queue coming from the previous iterations~\cite{he2020momentum}. 
Following the setting in MOCO~\cite{he2020momentum}, the branch of keys is the momentum-updated one of the branch of queries.

\myparagraph{Total Optimization Objectives.} 
For the contrastive loss, InfoNCE~\cite{oord2018representation} is employed to pull $q$ close to $k_{+}$ while pushing it away from other negative keys:
\begin{equation}
\small
\mathcal{L}_q = -\log \frac{\exp(q {\cdot} k_{+} / \tau)} {\exp(q {\cdot} k_{+} / \tau) + \sum_{k_-} \exp(q {\cdot} k_-  / \tau)},
\label{eq:contra_loss}
\end{equation}
where $\tau$ denotes a temperature hyper-parameter as in \cite{wu2018unsupervised}. The overall optimization function is written as:
\begin{equation}
\label{eq:pretraining_loss}
\small
\mathcal{L} = \mathcal{L}_q + \alpha\mathcal{L}_{e},
\end{equation}
where a Lagrange multiplier $\alpha$ is a fixed value that determines the trade-off between entropy and semantic integrity. 
Note that, the added additive noise is only a transitional component for entropy estimation in the omnipotent feature learning stage, and is discarded in the next two steps, \ie omnipotent feature compression and deployment.

\vspace{-1mm}
\subsection{Stage 2: Learning-based Feature Compression} \label{sec: feature compression}
\vspace{-1mm}

Similar to lossy image compression, the goal of lossy feature compression is simultaneously minimizing the size of bitstream and the distortion between $f$ and $\hat{f}$.
Such objectives can be formulated as minimizing $R+{\lambda}D_C$ (here we use $D_C$ to distinguish the $D$ in IF module), where the Lagrange multiplier $\lambda$ controls the trade-off between the rate $R$ and the distortion $D_C$ in feature level. $R$ denotes the rate of compressed feature and $D_C$ represents the distortion between $f$ and $\hat{f}$. Since quantization is non-differentiable, the additive uniform noise~\cite{balle2017end} is added to the latent variables during training for approximately rate estimation, which alters quantization to be differentiable. 
And, after quantization, the entropy coding is performed on latent variables $y$ to encode it into bitstream losslessly. Entropy coding here can be Huffman coding or arithmetic coding. Finally, for the omnipotent feature reconstruction, the decoder tend to reconstruct omnipotent features from $\hat{y}$. The R-D (rate-distortion) loss function can be written as:
\begin{equation}\label{eq:rd_loss}
\small
\mathcal{L}_{rd} = 
\mathop{\mathbb{E}}[-\log_{2}(p_{{\hat{\boldsymbol{y}}}|\boldsymbol{\psi}}({\hat{\boldsymbol{y}}}|\boldsymbol{\psi} ))]
+ 
{\lambda}\frac{1}{WH}\sum_{x=1}^{W}\sum_{y=1}^{H}(f_{x,y}-\hat{f}_{x,y})^2,
\end{equation}
where $W$ and $H$ denotes the width and height of features.

Moreover, since the features are compressed to handle downstream tasks better, we further protect its semantic fidelity in a deeper feature level. Particularly, the omnipotent feature $f$ and its reconstructed one $\hat{f}$ are passed through the backbone tail in the omnipotent feature learning stage, \ie layer2$\sim$layer4 in a normal ResNet. And then, the Euclidean distance is calculated between those two deeper feature representations of $f$ and $\hat{f}$ to construct this loss:
\begin{equation}
\label{eq:compression_loss_feat}
\small
\mathcal{L}_f = \sum_{i=2}^{4}\lambda_{i}\frac{1}{W_iH_i}\sum_{x=1}^{W_i}\sum_{y=1}^{H_i}(\phi_if_{x,y}-\phi_i\hat{f}_{x,y})^2,
\end{equation}
where $W_i$ and $H_i$ are widths and heights of feature maps, $\phi_i$ means a differentiable function, hyperparameter $\lambda_i$ controls the importances of distortions in different depths.
The overall loss function of feature compression is given by:
\begin{equation}
\small
\mathcal{L}_{com} = \mathcal{L}_{rd} + \mathcal{L}_f.
\end{equation}

Practically, we design the neural network for omnipotent feature compression, which is derived from the Mean \& Scale (M\&S) Hyperprior model~\cite{minnen2018joint}, and discretized Gaussian Mixture Likelihoods (GMM) entropy model~\cite{cheng2020learned}.

% Note that there are two autoencoders in our pipeline, however, with different architectures, implementation approaches, and functionality. 
% The former is for optimized with contrastive loss under entropy constraint, in order to filter out redundant information for analysis, without real quantization operation.
% While the latter is optimized for feature compression under the rate-distortion constraint, where quantization would be performed in practice. We report the detailed architectures of them in Supplementary.

Last but not least, there are two autoencoders in our pipeline, however, with different architectures, implementations, and functions. The first autoencoder in IF module is optimized with both contrastive loss and entropy constraint, without hard quantization operation in practice, acting as an information filter. The other autoencoder is used for feature compression, with hard quantization in practice. Detailed architectures are reported in \textbf{Supplementary}.

\vspace{-5mm}
\subsection{Stage 3: Feature Deployment and Task Supporting}\label{section: downstream supporting}
\vspace{-3mm}

After the omnipotent feature learning, the source data for machines has changed from images to omnipotent features. Therefore, the task models are trained with the learned omnipotent features $f$ and are evaluated with the reconstructed omnipotent features $\hat{f}$, to finally support the AI tasks.
% In practical, to take advantage of strong representation learning ability of contrastive learning, we take the pre-trained weights of backbone tail in omnipotent feature learning process as initialization for fine-tuning in downstream tasks.
% In practice, to take advantage of strong representation learning ability of contrastive learning, we take the pre-trained weights of backbone tail in omnipotent feature learning process as initialization for fine-tuning in downstream tasks.
Formally, only the backbone tail is fine-tuned for downstream tasks supporting, and the weights obtained in the omnipotent feature learning stage are used for a better initialization.

% The metric of object detection and semantic segmentation are the COCO-style AP and mean IoU (mIoU), respectively

\vspace{-4mm}

\section{Experiments}
\vspace{-1mm}
\subsection{Datasets}
\vspace{-1mm}
The training for both omnipotent feature learning and feature compression is conducted on the training set of the ImageNet~\cite{deng2009imagenet} dataset, which contains $\sim1.28$ million images of $1000$ classes. 
After the training of feature extraction and compression, we evaluate the transferability of the learned omnipotent features to downstream tasks on PASCAL VOC~\cite{everingham2010PASCAL}, MS COCO~\cite{lin2014microsoft} and Cityscapes~\cite{cordts2016cityscapes}. 
PASCAL VOC and MS COCO are the widely-used datasets for dense prediction tasks, \egno, object detection, instance segmentation.
Compared with PASCAL VOC, MS COCO
is larger and more challenging (more complicated scenes, more objects per image, and more categories to be predicted).
Cityscapes is a fundamental and challenging dataset for semantic segmentation, which contains 5000 high-quality images with the pixel-level annotations (2975, 500, and 1525 for the training, validation, and test sets respectively).

\subsection{Implementation Details}
\vspace{-1mm}
\myparagraph{Omnipotent feature learning.} 
With ResNet-50~\cite{he2016deep} as the basic architecture, the IF module takes the output of backbone head as input to obtain the omnipotent feature. 
In the omnipotent feature learning stage, the momentum update from one encoder to another is set to $0.999$ and the dictionary size is set to $65536$. 
Temperature in Eq. (\ref{eq:contra_loss}) is set to $0.2$.
The data augmentation operations and the use of MLP projection head are same as the previous contrastive learning related works~\cite{chen2020improved,grill2020bootstrap,chen2020simple,chen2021exploring,he2020momentum}. More specifically, the augmentations are random crop, random color distortion, random greyscale, random horizontal flip and random Gaussian blur. Besides, we load the weights that pre-training $800$ epochs with MOCO-v2~\cite{he2020momentum} to initialize the backbone head and backbone tail, and then keep all parameters fixed except the IF module for a stable training at the first $10$ epochs. After that, all the parameters are optimized together for another $190$ epochs. 
We adopt SGD as the optimizer with weight decay and momentum set as $10^{-4}$ and $0.9$. The batch size is $256$ and the learning rate is $10^{-3}$. $\alpha$ in Eq. (\ref{eq:pretraining_loss}) is experimentally set to $0.1$.

\myparagraph{Omnipotent Feature Compression.} 
We train the omnipotent feature compressor model for $400,000$ iterations with batch size of $32$. We employ the Adam~\cite{loshchilov2017decoupled} optimizer, where the learning rate is set to be $5\times10^{-5}$. Data augmentation is $256\times256$ random cropping. $\lambda$ in Eq. (\ref{eq:rd_loss}) is set to $2048$, and $\lambda_2, \lambda_3, \lambda_4$ in Eq. (\ref{eq:compression_loss_feat}) are set to $512$, $256$, $125$ respectively. 
Feature codecs with different rates are obtained by multiplying  $\lambda$, $\lambda_2$, $\lambda_3$, and $\lambda_4$ by a same coefficient.

\vspace{-1mm}
\subsection{Effectiveness and Superiority of Omni-ICM}
\vspace{-1mm}
\myparagraph{Evaluation Protocol.}
We evaluate the generalization of omnipotent features on different fundamental intelligent tasks by fine-tuning the backbone tail. Challenging and popular datasets are adapted for different tasks, \ie VOC object detection, COCO object detection, COCO instance segmentation, COCO pose estimation, Cityscapes semantic segmentation, and Cityscapes panoptic segmentation.
Experiments for Cityscapes semantic segmentation are implemented in \cite{mmseg2020} and others are implemented in \cite{wu2019detectron2}.
To evaluate the rate-distortion performance, the rate is measured by the bits per pixel (bpp), which is calculated by dividing the size of the feature bitstream by the number of pixels in the original image, and the distortion here represents metrics of different AI tasks.

\myparagraph{Comparison Approaches.}
We mainly compare our Omni-ICM with the most advanced traditional codecs (HEVC~\cite{sullivan2012overview}, VVC~\cite{bross2021overview}) and a learning-based compression method\cite{cheng2020learned}.
% For tradtional codecs, \textit{YUV420} is choosed as the configuration. 
To ensure the fairness of comparison, we use the pre-trained model that has trained for $800$ epochs on ImageNet~\cite{chen2020improved} as the initial weights and fine-tunes it on each task to get the well-trained networks for comparison, which is consistent with the operations taken by the current SOTA representation learning method, MOCO~\cite{he2020momentum}.
Then during evaluation of compared approaches, reconstructed images are input into these networks to obtain the final results. 
Our method and the compared methods follow the same training schedule for fine-tuning downstream tasks. Besides, in order to better understand the results, we provide results with uncompressed images or features performing intelligent tasks, which can be seen as baselines. We also report down-stream task performances with supervised pre-training in \textbf{Supplementary}.

\begin{figure}
  \centerline{\includegraphics[width=1.0\linewidth]{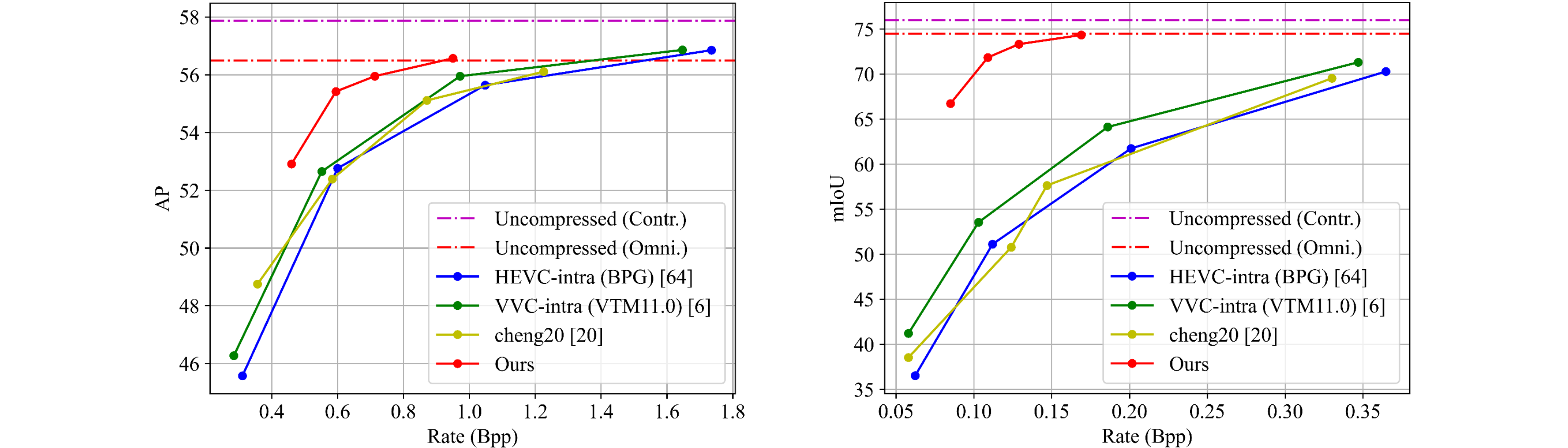}}
  \vspace{-4mm}
    \caption{\textbf{Object detection mAP on PASCAL VOC (left) and semantic segmentation mIoU on Cityscapes (right) under different bitrates.} We compare our method with two traditional codecs HEVC-intra~\cite{sullivan2012overview}, VVC-intra~\cite{bross2021overview}, and one learning-based codec~\cite{cheng2020learned}.
    % For object detection, the detectors (Faster R-CNN with C4-backbone) are trained on \texttt{trainval07+12} for $24$k iterations and evaluate on \texttt{test07}. The metric here is the COCO-style mAP. For semantic segmentation, an FCN-based structure is used. We train task networks on the \texttt{train\_fine} set which consist of 2975 images for $80$k iterations, and evaluate on the \texttt{val} set. The metric is mean IoU (mIoU), which is commonly used for semantic segmentation
    }
\label{fig:voc_det_cityscapes_segmentation}
\vspace{-6mm}
\end{figure}

\myparagraph{Object Detection on PASCAL VOC.}\label{section: voc_detection}
When evaluating on VOC object detection, we follow the common protocol that fine-tuning a Faster R-CNN detector (C4-backbone) on the VOC \texttt{trainval07+12} set and testing on the VOC \texttt{test2007} set.
The image scale is in $[640, 800]$ pixels during training and is $800$ at inference as default. 
Note that the image resolution has changed before inputting into the task model.
% , and the number of pixels on the image is about four times as much as before. 
For the fairness of comparison, we don't perform any resizing operations on the features, and we regard the original image as the source data to be compressed so that we calculate the rate by dividing the size of the bitstream file of feature by the number of pixels of the original image.
Other tasks that need resizing during preprocessing all obey this setting, \ie instance segmentation, pose estimation. Fig. \ref{fig:voc_det_cityscapes_segmentation} (left) shows the results of detection. Our method achieves the best performance (lower rate, higher precision).

\myparagraph{Semantic Segmentation on Cityscapes.}
For semantic segmentation, an FCN-based structure is used. 
We train task networks on the \texttt{train\_fine} set which consist of 2975 images for $80$k iterations, and evaluate on the \texttt{val} set.
Results are shown in Fig. \ref{fig:voc_det_cityscapes_segmentation} (right). Similarly, our method is also the best scheme.

\begin{figure}
  \centerline{\includegraphics[width=1.0\linewidth]{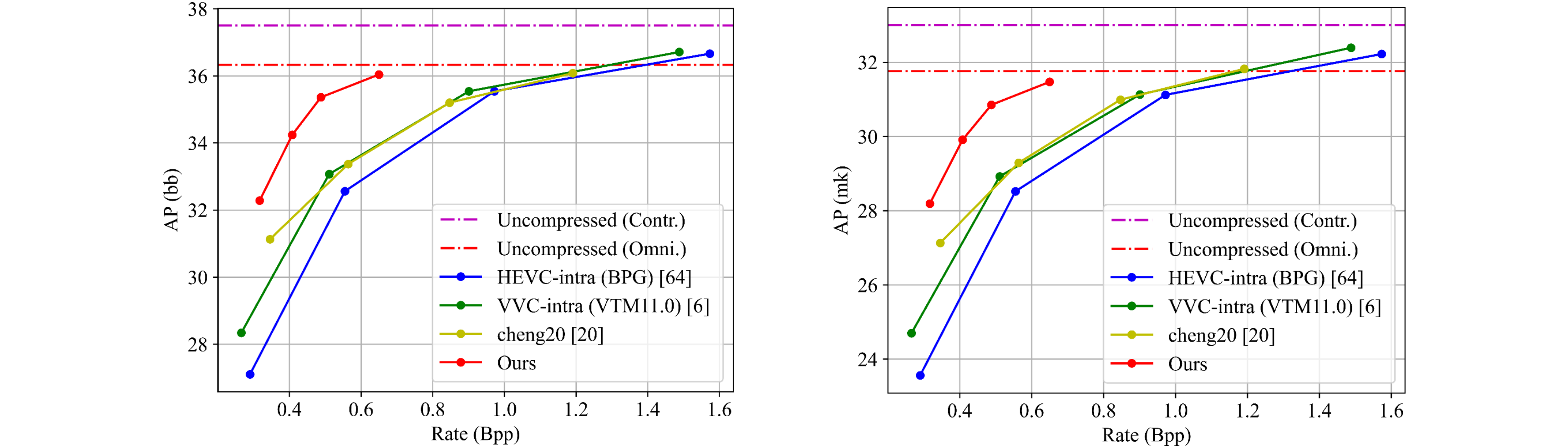}}
  \vspace{-4mm}
    \caption{\textbf{Object detecion and instance segmentation on MS COCO.} The metrics here include mean bounding box AP (AP$^{bb}$) and mask AP (AP$^{mk}$).
    % The detector (Mask R-CNN with C4-backbone) are trained on \texttt{train2017} with default $1\times$ schedule and evaluate on \texttt{val2017}. The metrics here include bounding box AP (AP$^{bb}$) and mask AP (AP$^{mk}$).
    }
\label{fig:feature_deployment_coco_instance}
\vspace{-4mm}
\end{figure}

\myparagraph{Object Detection and Instance Segmentation on MS COCO.} Following the setting in \cite{he2020momentum}, we evaluate object detection and instance segmentation by fine-tuning a Mask R-CNN detector (C4-backbone) on COCO \texttt{train2017} split with the standard $1\times$ schedule and evaluating on COCO \texttt{val2017} split, with BN tuned and synchronized across GPUs. The image scale is in [$640$, $800$] pixels during training and is $800$ at inference as default, same as that for PASCAL VOC. The comparison is shown in Fig. \ref{fig:feature_deployment_coco_instance}. Our method also achieves the best performance, and significantly outperforms the other codecs.

\begin{figure}
  \centerline{\includegraphics[width=1.0\linewidth]{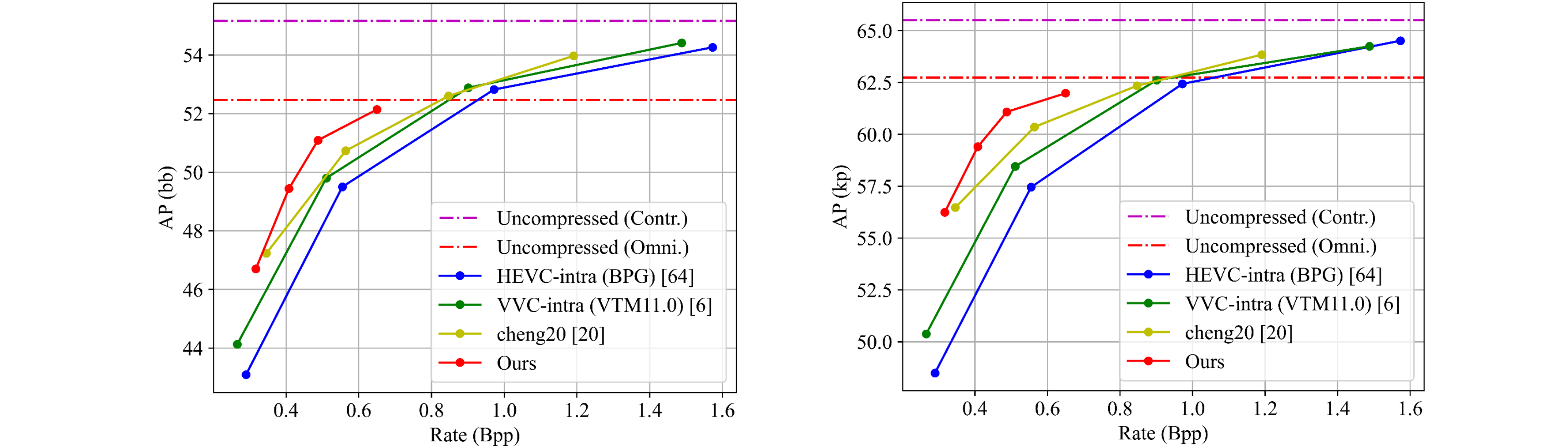}}
  \vspace{-4mm}
    \caption{\textbf{Pose estimation on MS COCO.} 
    % Mask R-CNN (with R50-FPN) is fine-tuned on COCO \texttt{train2017} and evaluated on \texttt{val2017}. The schedule is $1\times$. 
    Results of person detection (AP$^{bb}$) and keypoint detection (AP$^{kp}$) are illustrated. 
    }
\label{fig:feature_deployment_coco_pose}
\vspace{-4mm}
\end{figure}

\begin{figure}
  \centerline{\includegraphics[width=1.0\linewidth]{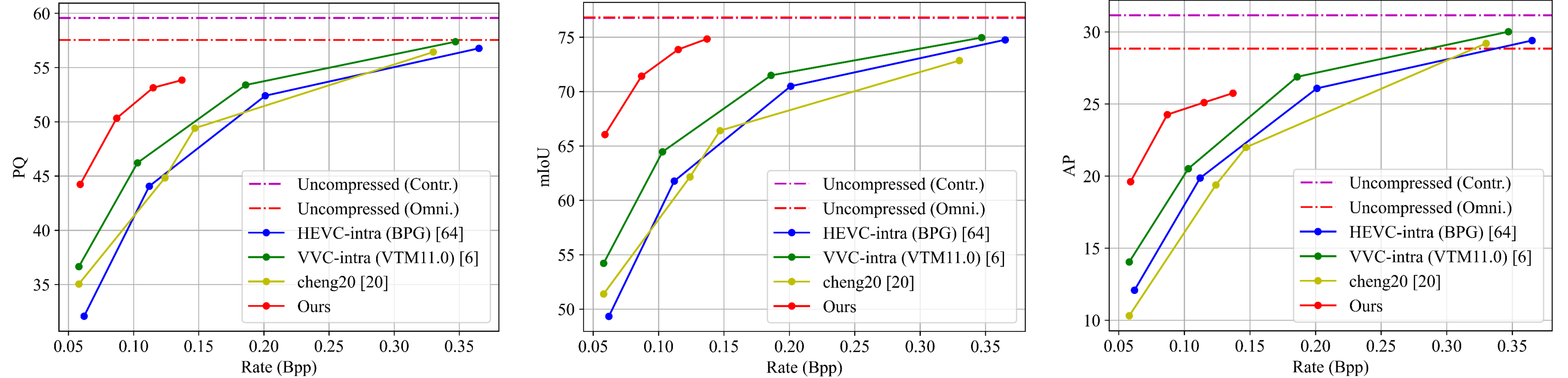}}
  \vspace{-4mm}
    \caption{\textbf{Panoptic segmentation on Cityscapes.} 
    % Panoptic-deeplab~\cite{cheng2020panoptic} is used. We train task networks on the \texttt{train\_fine} set for 90k iterations, and evaluate on the \texttt{val} set. 
    % We report results of PQ, mIoU, and AP in this figure. 
    PQ, mIoU, and AP are reported. PQ is the metric of panoptic segmentation which measures the performance for both stuff and things in a uniform manner, mIoU is the metric of semantic segmentation, and AP is the metric of instance segmentation. 
    }
\label{fig:feature_deployment_panoptic}
\vspace{-4mm}
\end{figure}

\myparagraph{More Downstream Tasks.}
% Figure \ref{fig:voc_det_cityscapes_segmentation} (right), \ref{fig:feature_deployment_coco_pose}, \ref{fig:feature_deployment_panoptic} show results on more downstream tasks:
Fig. \ref{fig:feature_deployment_coco_pose}, \ref{fig:feature_deployment_panoptic} show results on more downstream tasks:
% (details in \textbf{Supplementary}):

% \textit{Cityscapes semantic segmentation}:
% As shown in Figure \ref{fig:voc_det_cityscapes_segmentation} (right), our Omni-ICM outperforms other methods by a large margin.

\textit{COCO pose estimation}:
Mask R-CNN (with R50-FPN) is fine-tuned on COCO \texttt{train2017} and evaluated on \texttt{val2017}. The schedule is $1\times$.

Results are illustrated in Fig. \ref{fig:feature_deployment_coco_pose}. 
Although Omni-ICM is better than other methods, however, there exists an obvious gap (more than 2 points in both person detection and keypoint detection) between the best performance at high bitrate. This also indicates the superiority of our method at lower bitrates.

\textit{Cityscapes panoptic segmentation}~\cite{kirillov2019panoptic,cheng2020panoptic}:
Panoptic-deeplab~\cite{cheng2020panoptic} is used for this task. We train task networks on the \texttt{train\_fine} set for 90k iterations, and evaluate on the \texttt{val} set. 
Results of PQ, mIoU, and AP are reported for panoptic segmentation in Fig. \ref{fig:feature_deployment_panoptic}. 
% PQ is the metric of panoptic segmentation which measures the performance for both stuff and things in a uniform manner, mIoU is the metric of semantic segmentation, and AP is the metric of instance segmentation. 
The performance of mIoU is similar to Fig. \ref{fig:voc_det_cityscapes_segmentation} (right). We can observe that our method achieves the better R-D performance, which means it can use less bits to achieve higher task performance.

% However, significant gaps between the best performances at high bitrate for both PQ and AP are also observed, although the R-D performance is better.

\myparagraph{Discussion.}
For the case of image coding for machines (ICM), Omni-ICM outperforms the most advanced hand-craft traditional codecs and a learning-based codec by remarkable margins on 6 fundamental intelligent tasks. Besides, we also observe some hidden limitations.
Results in Fig. \ref{fig:feature_deployment_coco_pose} and Fig. \ref{fig:feature_deployment_panoptic}
show the potential performance gaps at the highest bitrate.
We speculate that this is caused by two reasons.
The first one is the discrepancy between datasets, the ImageNet is mainly composed of images with a single conspicuous target in natural scenes, while the number of targets in MS COCO and Cityscapes is diversified, and the scales of targets are also various.
The second reason is that training by instance discrimination~\cite{he2020momentum,chen2020simple,oord2018representation} forces the model to focus more on the conspicuous part of the image, which is not conducive to the preservation of local semantic information that occurs frequently in the above two datasets.

% We leave the further improvement in future work.

\vspace{-1mm}
\subsection{Comparison with SOTA ICM-related methods}
\vspace{-1mm}

First, we must emphasize that our method focuses on a new ICM paradigm of ``one bitstream covers multiple different tasks'', and we are also the first to report results on such wide range of intelligent tasks and widely accepted datasets. To our knowledge, there is currently no similar work has studied such general problem as we did. Recent ICM-related methods, \egno, the traditional codec based RoI bit location scheme~\cite{huang2021visual} and the learning based joint training codec~\cite{le2021image}, mostly only focus on specific AI tasks and report the corresponding results, which makes it hard to directly compare these methods with
ours and guarantees fairness. Besides, most of these methods
didn’t release codes, which further makes the fair comparison become more difficult. Despite this, we still reproduced two SOTA ICM-related competitors~\cite{huang2021visual,le2021image} following their
papers. The RoI based~\cite{huang2021visual} is optimized and evaluated for every task. The end-to-end joint training based~\cite{le2021image} is trained with object detetion on PASCAL VOC dataset and evaluated on all tasks. Table~\ref{tab: bd-rate} shows the comparison, our method significantly outperforms them by a large margin (the lower the better).

\begin{table}[htbp]
\centering
\footnotesize
\vspace{-1mm}
\caption{
Comparison with two SOTA ICM methods: RoI based bit allocation (RoI)~\cite{huang2021visual} and task-driven joint training (Joint)~\cite{le2021image}. Three AI tasks of detection (Det.), instance segmentation (Ins.), and semantic segmentation (Sem.) are used for evaluation. Bjøntegaard Delta rate (BD-rate) saving w.r.t the AI task performances is taken as metric (more lower, more better). 
Three mainstream codecs are taken as anchors: HEVC~\cite{sullivan2012overview}, VVC~\cite{bross2021overview}, a SOTA learned based codec~\cite{cheng2020learned} (noted as cheng). HEVC is taken as the benchmark.
Note that, the results with ``*'' are converted from the original paper. The best performance of each task is marked in bold.
% 1). H denotes HEVC, 2). V denotes VVC, 3). Che is a learning-based codec [19]. 
% Note that, \tcr{`Red'} results are reported from the original paper [32], our re-implementation performs better than their reports, but is still inferior to our method.
}
\vspace{-2.5mm}
\setlength{\tabcolsep}{0.8mm}{
\begin{tabular}{ccccccc}
\hline
Datasets            & HEVC+RoI~\cite{huang2021visual} & VVC    & VVC+RoI~\cite{huang2021visual} & cheng & cheng+Joint~\cite{le2021image} & Ours   \\ \hline
Det. (VOC)    & -17.6   & -9.0  & -32.9*  & -0.9  & -14.4        & \textbf{-35.1} \\
Det. (COCO)   & -17.2   & -14.4 & -32.4  & -10.7  & -3.0         & \textbf{-43.9} \\
Ins. (COCO)   & -11.9   & -14.1 & -39.1*  & -12.8  & -5.4         & \textbf{-42.8} \\
Sem. (City.) & 4.3     & -24.6 & -25.4  & -0.8   & 2.3          & \textbf{-72.0} \\ \hline
\end{tabular}}
\label{tab: bd-rate}
\vspace{-5mm}
\end{table}

\vspace{-1mm}
\subsection{Ablation Study}
\vspace{-1mm}
We implement ablation studies by pre-training on ImageNet and fine-tuning on VOC0712 object detection, as introduced in \ref{section: voc_detection}.

\begin{figure}
  \centerline{\includegraphics[width=1.0\linewidth]{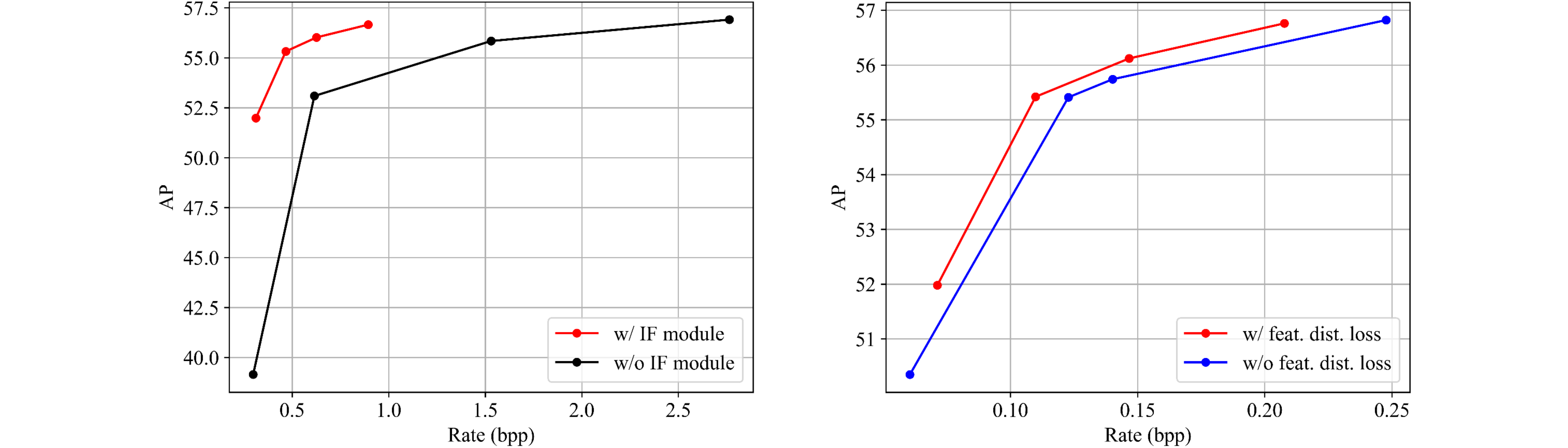}}
  \vspace{-4mm}
    \caption{Ablation studies on \textbf{IF module} (left) and \textbf{feature level distortion loss} (right), respectively.}
\label{fig:ablation}
\vspace{-6mm}
\end{figure}

\myparagraph{Study on IF module.} The first graph in Fig. \ref{fig:ablation} illustrates the result that validate the contribution of IF module.
For the case without IF module, the features output by layer1 of the ResNet-50 network pre-trained by contrastive learning are employed for task supporting and compression.
Thus, we fix parameters in stem layer and layer1, and then fine-tunes the task model on PASCAL VOC detection. 
A feature codec with the same architecture and training schedule as that in Section \ref{sec: feature compression} is trained for feature compression.
As we can see, in the absence of IF module, compressing features directly can achieve satisfying performance with low coding efficiency. However, our Omni-ICM can achieve comparable performance with much lower bitrate.

\myparagraph{Feature level distortion loss.} The second graph in Fig. \ref{fig:ablation} presents the ablation study about feature-level distortion in Eq. (\ref{eq:compression_loss_feat}). It indicates that the loss of feature level distortion helps protect semantic information.

\vspace{-4mm}

\vspace{-6mm}

\begin{figure}
  \centerline{\includegraphics[width=1.0\linewidth]{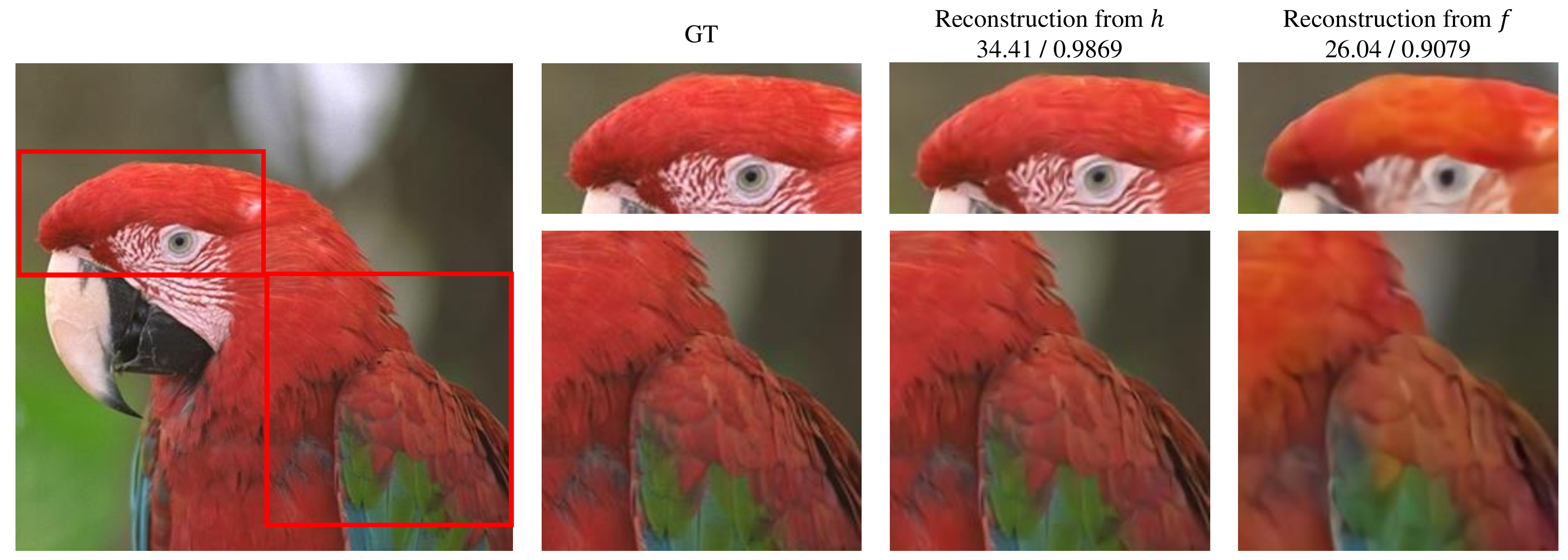}}
  \vspace{-4mm}
    \caption{\textbf{Reconstruction of features before and after IF module.} The numbers on the top of the crop images indicate PSNR (dB) / MS-SSIM of an entire image.}
\label{fig:qualitative_kodak23}
\vspace{-6mm}
\end{figure}

\vspace{-6mm}

\begin{figure}
  \centerline{\includegraphics[width=1.0\linewidth]{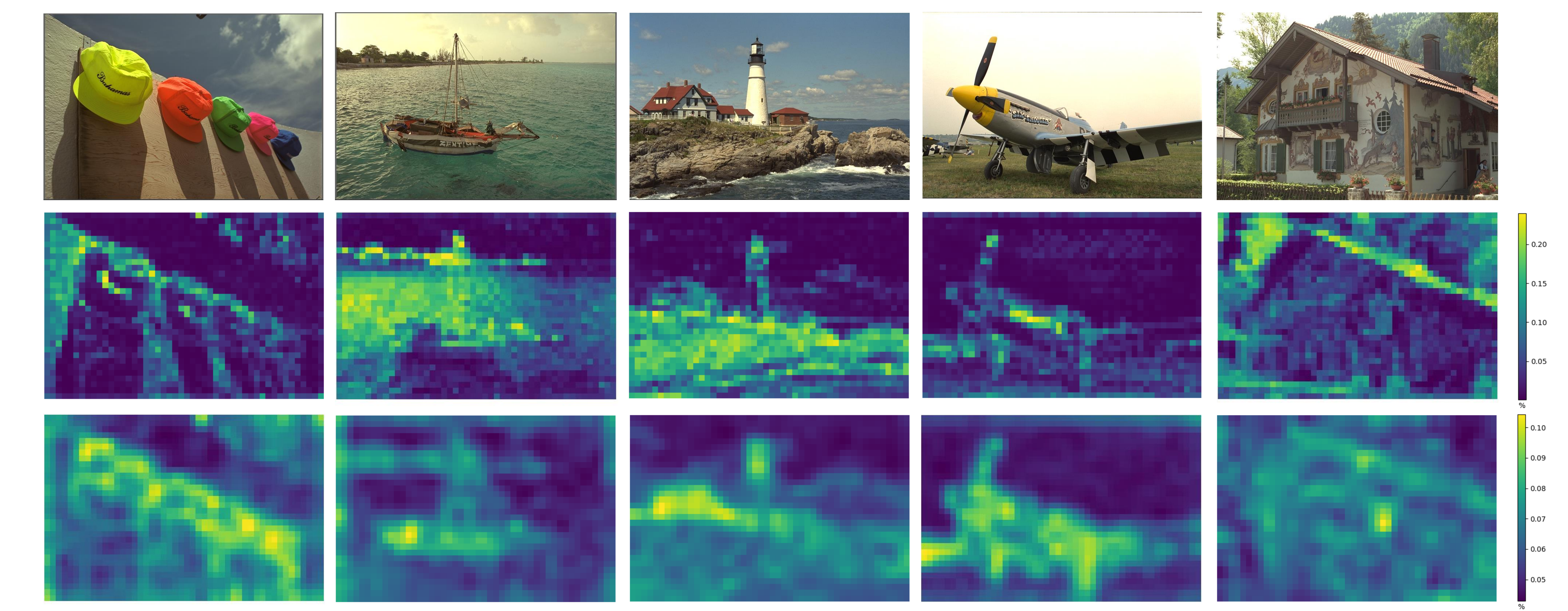}}
  \vspace{-4mm}
    \caption{\textbf{The bit allocation maps in learning-based codec~\cite{cheng2020learned} (second line) and  our IF module (third line), respectively.} The first line is ground truth.}
\label{fig:qualitative_bit_allocation}
\vspace{-4mm}
\end{figure}

% \vspace{-4mm}
\subsection{Vision Analysis and Insights}
\vspace{-1mm}
\myparagraph{Reconstruction Results.}\label{section:vision_analysis - reconstruction} To better understand the functionability of the IF module, we additionally train two decoders to visualize the reconstruction results of features before and after IF module, \ie $h$ and $f$. Both decoders are optimized with MSE loss.
As illustrated in Fig. \ref{fig:qualitative_kodak23}, images reconstructed from $h$ contain slight color difference, and textures are relatively complete. But images reconstructed from $f$ suffer obvious color difference and texture loss. 
It can be observed that IF module drops out some color information and detailed texture information that has a slight influence on intelligent analytics.
We report details for training these two decoders in \textbf{Supplementary}.

\myparagraph{Bit allocation Map.} As is illustrated in Fig.~\ref{fig:qualitative_bit_allocation}, we also visualize the bit allocation maps in IF module and that in the learning-based codec~\cite{cheng2020learned} optimized with MSE loss. Learning-based codec tends to focus on areas with large, irregular, and complex textures, \egno, walls, water surfaces, rocks, and eaves.
But our IF module pays less attention to the texture details in the image and more attention to the objects, which is crucial for the understanding of images.

\section{Conclusion}
\vspace{-1mm}
We presented a novel framework for image coding for machines (Omni-ICM) based on extracting and compressing a general and compact feature, dubbed omnipotent feature. 
The omnipotent feature is learned by elegantly combining the contrastive learning and entropy constraint through a new IF module, which coordinates semantics modeling and redundancy removing in our framework by adaptively filtering information that weakly related to AI tasks. Extensive experiments show an outstanding performance of our proposed Omni-ICM framework compared to the SOTA traditional  and learning-based approaches.

\section*{Acknowledgement}
This work was supported in part by NSFC under Grant U1908209, 62021001 and the National Key Research and Development Program of China 2018AAA0101400.

% \clearpage

\bibliographystyle{splncs04}
% \bibliographystyle{unsrt} 
% \bibliography{egbib}
% \bibliographystyle{ieee_fullname}
\bibliography{reference}

\begin{thebibliography}{10}
\providecommand{\url}[1]{\texttt{#1}}
\providecommand{\urlprefix}{URL }
\providecommand{\doi}[1]{https://doi.org/#1}

\bibitem{alemi2016deep}
Alemi, A.A., Fischer, I., Dillon, J.V., Murphy, K.: Deep variational
  information bottleneck. arXiv preprint arXiv:1612.00410  (2016)

\bibitem{badrinarayanan2017segnet}
Badrinarayanan, V., Kendall, A., Cipolla, R.: Segnet: A deep convolutional
  encoder-decoder architecture for image segmentation. TPAMI  \textbf{39}(12),
  2481--2495 (2017)

\bibitem{bajic2021collaborative}
Baji{\'c}, I.V., Lin, W., Tian, Y.: Collaborative intelligence: Challenges and
  opportunities. In: ICASSP 2021-2021 IEEE International Conference on
  Acoustics, Speech and Signal Processing (ICASSP). pp. 8493--8497. IEEE (2021)

\bibitem{balle2017end}
Ball{\'e}, J., Laparra, V., Simoncelli, E.P.: End-to-end optimized image
  compression. In: ICLR (2017)

\bibitem{balle2018variational}
Ball{\'e}, J., Minnen, D., Singh, S., Hwang, S.J., Johnston, N.: Variational
  image compression with a scale hyperprior. In: ICLR (2018)

\bibitem{blau2018perception}
Blau, Y., Michaeli, T.: The perception-distortion tradeoff. In: CVPR. pp.
  6228--6237 (2018)

\bibitem{bolya2019yolact}
Bolya, D., Zhou, C., Xiao, F., Lee, Y.J.: Yolact: Real-time instance
  segmentation. In: ICCV. pp. 9157--9166 (2019)

\bibitem{bross2021overview}
Bross, B., Wang, Y.K., Ye, Y., Liu, S., Chen, J., Sullivan, G.J., Ohm, J.R.:
  Overview of the versatile video coding (vvc) standard and its applications.
  TCSVT  (2021)

\bibitem{cai2021novel}
Cai, Q., Chen, Z., Wu, D., Liu, S., Li, X.: A novel video coding strategy in
  hevc for object detection. TCSVT  (2021)

\bibitem{caron2020unsupervised}
Caron, M., Misra, I., Mairal, J., Goyal, P., Bojanowski, P., Joulin, A.:
  Unsupervised learning of visual features by contrasting cluster assignments.
  arXiv preprint arXiv:2006.09882  (2020)

\bibitem{chen2017deeplab}
Chen, L.C., Papandreou, G., Kokkinos, I., Murphy, K., Yuille, A.L.: Deeplab:
  Semantic image segmentation with deep convolutional nets, atrous convolution,
  and fully connected crfs. TPAMI  \textbf{40}(4),  834--848 (2017)

\bibitem{chen2018encoder}
Chen, L.C., Zhu, Y., Papandreou, G., Schroff, F., Adam, H.: Encoder-decoder
  with atrous separable convolution for semantic image segmentation. In: ECCV.
  pp. 801--818 (2018)

\bibitem{chen2020simple}
Chen, T., Kornblith, S., Norouzi, M., Hinton, G.: A simple framework for
  contrastive learning of visual representations. In: ICML. pp. 1597--1607.
  PMLR (2020)

\bibitem{chen2020improved}
Chen, X., Fan, H., Girshick, R., He, K.: Improved baselines with momentum
  contrastive learning. arXiv preprint arXiv:2003.04297  (2020)

\bibitem{chen2021exploring}
Chen, X., He, K.: Exploring simple siamese representation learning. In: CVPR.
  pp. 15750--15758 (2021)

\bibitem{chen2019learning}
Chen, Z., He, T., Jin, X., Wu, F.: Learning for video compression. IEEE
  Transactions on Circuits and Systems for Video Technology  \textbf{30}(2),
  566--576 (2019)

\bibitem{chen2020data}
Chen, Z., Duan, L.Y., Wang, S., Lin, W., Kot, A.C.: Data representation in
  hybrid coding framework for feature maps compression. In: 2020 IEEE
  International Conference on Image Processing (ICIP). pp. 3094--3098. IEEE
  (2020)

\bibitem{chen2019lossy}
Chen, Z., Fan, K., Wang, S., Duan, L.Y., Lin, W., Kot, A.: Lossy intermediate
  deep learning feature compression and evaluation. In: ACM MM. pp. 2414--2422
  (2019)

\bibitem{chen2019toward}
Chen, Z., Fan, K., Wang, S., Duan, L., Lin, W., Kot, A.C.: Toward intelligent
  sensing: Intermediate deep feature compression. TIP  \textbf{29},  2230--2243
  (2019)

\bibitem{chen2018intermediate}
Chen, Z., Lin, W., Wang, S., Duan, L., Kot, A.C.: Intermediate deep feature
  compression: the next battlefield of intelligent sensing. arXiv preprint
  arXiv:1809.06196  (2018)

\bibitem{cheng2020panoptic}
Cheng, B., Collins, M.D., Zhu, Y., Liu, T., Huang, T.S., Adam, H., Chen, L.C.:
  Panoptic-deeplab: A simple, strong, and fast baseline for bottom-up panoptic
  segmentation. In: CVPR. pp. 12475--12485 (2020)

\bibitem{cheng2020learned}
Cheng, Z., Sun, H., Takeuchi, M., Katto, J.: Learned image compression with
  discretized gaussian mixture likelihoods and attention modules. In: CVPR. pp.
  7939--7948 (2020)

\bibitem{choi2018high}
Choi, H., Bajic, I.V.: High efficiency compression for object detection. In:
  2018 IEEE International Conference on Acoustics, Speech and Signal Processing
  (ICASSP). pp. 1792--1796. IEEE (2018)

\bibitem{codevilla2021learned}
Codevilla, F., Simard, J.G., Goroshin, R., Pal, C.: Learned image compression
  for machine perception. arXiv preprint arXiv:2111.02249  (2021)

\bibitem{mmseg2020}
Contributors, M.: {MMSegmentation}: Openmmlab semantic segmentation toolbox and
  benchmark. \url{https://github.com/open-mmlab/mmsegmentation} (2020)

\bibitem{cordts2016cityscapes}
Cordts, M., Omran, M., Ramos, S., Rehfeld, T., Enzweiler, M., Benenson, R.,
  Franke, U., Roth, S., Schiele, B.: The cityscapes dataset for semantic urban
  scene understanding. In: CVPR. pp. 3213--3223 (2016)

\bibitem{deng2009imagenet}
Deng, J., Dong, W., Socher, R., Li, L.J., Li, K., Fei-Fei, L.: Imagenet: A
  large-scale hierarchical image database. In: CVPR. pp. 248--255. Ieee (2009)

\bibitem{duan2020video}
Duan, L., Liu, J., Yang, W., Huang, T., Gao, W.: Video coding for machines: A
  paradigm of collaborative compression and intelligent analytics. TIP
  \textbf{29},  8680--8695 (2020)

\bibitem{everingham2010PASCAL}
Everingham, M., Van~Gool, L., Williams, C.K., Winn, J., Zisserman, A.: The
  pascal visual object classes (voc) challenge. IJCV  \textbf{88}(2),  303--338
  (2010)

\bibitem{finn2017model}
Finn, C., Abbeel, P., Levine, S.: Model-agnostic meta-learning for fast
  adaptation of deep networks. In: International conference on machine
  learning. pp. 1126--1135. PMLR (2017)

\bibitem{goodfellow2014generative}
Goodfellow, I., Pouget-Abadie, J., Mirza, M., Xu, B., Warde-Farley, D., Ozair,
  S., Courville, A., Bengio, Y.: Generative adversarial nets. NeurIPS
  \textbf{27} (2014)

\bibitem{grill2020bootstrap}
Grill, J.B., Strub, F., Altch{\'e}, F., Tallec, C., Richemond, P.H.,
  Buchatskaya, E., Doersch, C., Pires, B.A., Guo, Z.D., Azar, M.G., et~al.:
  Bootstrap your own latent: A new approach to self-supervised learning. arXiv
  preprint arXiv:2006.07733  (2020)

\bibitem{guo2021causal}
Guo, Z., Zhang, Z., Feng, R., Chen, Z.: Causal contextual prediction for
  learned image compression. IEEE Transactions on Circuits and Systems for
  Video Technology  \textbf{32}(4),  2329--2341 (2021)

\bibitem{guo2021soft}
Guo, Z., Zhang, Z., Feng, R., Chen, Z.: Soft then hard: Rethinking the
  quantization in neural image compression. In: International Conference on
  Machine Learning. pp. 3920--3929. PMLR (2021)

\bibitem{he2020momentum}
He, K., Fan, H., Wu, Y., Xie, S., Girshick, R.: Momentum contrast for
  unsupervised visual representation learning. In: CVPR. pp. 9729--9738 (2020)

\bibitem{he2017mask}
He, K., Gkioxari, G., Doll{\'a}r, P., Girshick, R.: Mask r-cnn. In: ICCV. pp.
  2961--2969 (2017)

\bibitem{he2016deep}
He, K., Zhang, X., Ren, S., Sun, J.: Deep residual learning for image
  recognition. In: CVPR. pp. 770--778 (2016)

\bibitem{he2019beyond}
He, T., Sun, S., Guo, Z., Chen, Z.: Beyond coding: Detection-driven image
  compression with semantically structured bit-stream. In: 2019 Picture Coding
  Symposium (PCS). pp.~1--5. IEEE (2019)

\bibitem{hore2010image}
Hore, A., Ziou, D.: Image quality metrics: Psnr vs. ssim. In: 2010 20th
  international conference on pattern recognition. pp. 2366--2369. IEEE (2010)

\bibitem{hospedales2020meta}
Hospedales, T., Antoniou, A., Micaelli, P., Storkey, A.: Meta-learning in
  neural networks: A survey. arXiv preprint arXiv:2004.05439  (2020)

\bibitem{hu2020towards}
Hu, Y., Yang, S., Yang, W., Duan, L.Y., Liu, J.: Towards coding for human and
  machine vision: A scalable image coding approach. In: 2020 IEEE International
  Conference on Multimedia and Expo (ICME). pp.~1--6. IEEE (2020)

\bibitem{huang2021visual}
Huang, Z., Jia, C., Wang, S., Ma, S.: Visual analysis motivated rate-distortion
  model for image coding. In: 2021 IEEE International Conference on Multimedia
  and Expo (ICME). pp.~1--6. IEEE (2021)

\bibitem{jing2020self}
Jing, L., Tian, Y.: Self-supervised visual feature learning with deep neural
  networks: A survey. TPAMI  (2020)

\bibitem{johnston2018improved}
Johnston, N., Vincent, D., Minnen, D., Covell, M., Singh, S., Chinen, T.,
  Hwang, S.J., Shor, J., Toderici, G.: Improved lossy image compression with
  priming and spatially adaptive bit rates for recurrent networks. In: CVPR.
  pp. 4385--4393 (2018)

\bibitem{kim2021joint}
Kim, J.H., Heo, B., Lee, J.S.: Joint global and local hierarchical priors for
  learned image compression. arXiv preprint arXiv:2112.04487  (2021)

\bibitem{kingma2014adam}
Kingma, D.P., Ba, J.: Adam: A method for stochastic optimization. arXiv
  preprint arXiv:1412.6980  (2014)

\bibitem{kirillov2019panoptic}
Kirillov, A., He, K., Girshick, R., Rother, C., Doll{\'a}r, P.: Panoptic
  segmentation. In: CVPR. pp. 9404--9413 (2019)

\bibitem{le2021image}
Le, N., Zhang, H., Cricri, F., Ghaznavi-Youvalari, R., Rahtu, E.: Image coding
  for machines: An end-to-end learned approach. In: ICASSP 2021-2021 IEEE
  International Conference on Acoustics, Speech and Signal Processing (ICASSP).
  pp. 1590--1594. IEEE (2021)

\bibitem{li2020learning}
Li, M., Zuo, W., Gu, S., You, J., Zhang, D.: Learning content-weighted deep
  image compression. TPAMI  (2020)

\bibitem{li2018learning}
Li, M., Zuo, W., Gu, S., Zhao, D., Zhang, D.: Learning convolutional networks
  for content-weighted image compression. In: CVPR. pp. 3214--3223 (2018)

\bibitem{li2021task}
Li, X., Shi, J., Chen, Z.: Task-driven semantic coding via reinforcement
  learning. arXiv preprint arXiv:2106.03511  (2021)

\bibitem{lin2017feature}
Lin, T.Y., Doll{\'a}r, P., Girshick, R., He, K., Hariharan, B., Belongie, S.:
  Feature pyramid networks for object detection. In: CVPR. pp. 2117--2125
  (2017)

\bibitem{lin2014microsoft}
Lin, T.Y., Maire, M., Belongie, S., Hays, J., Perona, P., Ramanan, D.,
  Doll{\'a}r, P., Zitnick, C.L.: Microsoft coco: Common objects in context. In:
  ECCV. pp. 740--755. Springer (2014)

\bibitem{liu2018path}
Liu, S., Qi, L., Qin, H., Shi, J., Jia, J.: Path aggregation network for
  instance segmentation. In: CVPR. pp. 8759--8768 (2018)

\bibitem{long2015fully}
Long, J., Shelhamer, E., Darrell, T.: Fully convolutional networks for semantic
  segmentation. In: CVPR. pp. 3431--3440 (2015)

\bibitem{loshchilov2017decoupled}
Loshchilov, I., Hutter, F.: Decoupled weight decay regularization. arXiv
  preprint arXiv:1711.05101  (2017)

\bibitem{mentzer2018conditional}
Mentzer, F., Agustsson, E., Tschannen, M., Timofte, R., Van~Gool, L.:
  Conditional probability models for deep image compression. In: CVPR. pp.
  4394--4402 (2018)

\bibitem{mentzer2020high}
Mentzer, F., Toderici, G.D., Tschannen, M., Agustsson, E.: High-fidelity
  generative image compression. NeurIPS  \textbf{33},  11913--11924 (2020)

\bibitem{minnen2018joint}
Minnen, D., Ball{\'e}, J., Toderici, G.: Joint autoregressive and hierarchical
  priors for learned image compression. In: NeurIPS (2018)

\bibitem{minnen2020channel}
Minnen, D., Singh, S.: Channel-wise autoregressive entropy models for learned
  image compression. In: 2020 IEEE International Conference on Image Processing
  (ICIP). pp. 3339--3343. IEEE (2020)

\bibitem{newell2016stacked}
Newell, A., Yang, K., Deng, J.: Stacked hourglass networks for human pose
  estimation. In: ECCV. pp. 483--499. Springer (2016)

\bibitem{noroozi2016unsupervised}
Noroozi, M., Favaro, P.: Unsupervised learning of visual representations by
  solving jigsaw puzzles. In: ECCV. pp. 69--84. Springer (2016)

\bibitem{oord2018representation}
Oord, A.v.d., Li, Y., Vinyals, O.: Representation learning with contrastive
  predictive coding. arXiv preprint arXiv:1807.03748  (2018)

\bibitem{pathak2016context}
Pathak, D., Krahenbuhl, P., Donahue, J., Darrell, T., Efros, A.A.: Context
  encoders: Feature learning by inpainting. In: CVPR. pp. 2536--2544 (2016)

\bibitem{rabbani2002overview}
Rabbani, M., Joshi, R.: An overview of the jpeg 2000 still image compression
  standard. Signal processing: Image communication  \textbf{17}(1),  3--48
  (2002)

\bibitem{redmon2016you}
Redmon, J., Divvala, S., Girshick, R., Farhadi, A.: You only look once:
  Unified, real-time object detection. In: CVPR. pp. 779--788 (2016)

\bibitem{redmon2017yolo9000}
Redmon, J., Farhadi, A.: Yolo9000: better, faster, stronger. In: CVPR. pp.
  7263--7271 (2017)

\bibitem{ren2015faster}
Ren, S., He, K., Girshick, R., Sun, J.: Faster r-cnn: Towards real-time object
  detection with region proposal networks. NeurIPS  \textbf{28},  91--99 (2015)

\bibitem{singh2020end}
Singh, S., Abu-El-Haija, S., Johnston, N., Ball{\'e}, J., Shrivastava, A.,
  Toderici, G.: End-to-end learning of compressible features. In: 2020 IEEE
  International Conference on Image Processing (ICIP). pp. 3349--3353. IEEE
  (2020)

\bibitem{snell2017prototypical}
Snell, J., Swersky, K., Zemel, R.: Prototypical networks for few-shot learning.
  Advances in neural information processing systems  \textbf{30} (2017)

\bibitem{song2021variable}
Song, M., Choi, J., Han, B.: Variable-rate deep image compression through
  spatially-adaptive feature transform. In: ICCV. pp. 2380--2389 (2021)

\bibitem{sullivan2012overview}
Sullivan, G.J., Ohm, J.R., Han, W.J., Wiegand, T.: Overview of the high
  efficiency video coding (hevc) standard. TCSVT  \textbf{22}(12),  1649--1668
  (2012)

\bibitem{sun2020semantic}
Sun, S., He, T., Chen, Z.: Semantic structured image coding framework for
  multiple intelligent applications. TCSVT  (2020)

\bibitem{tishby2000information}
Tishby, N., Pereira, F.C., Bialek, W.: The information bottleneck method. arXiv
  preprint physics/0004057  (2000)

\bibitem{toderici2015variable}
Toderici, G., O'Malley, S.M., Hwang, S.J., Vincent, D., Minnen, D., Baluja, S.,
  Covell, M., Sukthankar, R.: Variable rate image compression with recurrent
  neural networks. arXiv preprint arXiv:1511.06085  (2015)

\bibitem{vanschoren2018meta}
Vanschoren, J.: Meta-learning: A survey. arXiv preprint arXiv:1810.03548
  (2018)

\bibitem{wallace1992jpeg}
Wallace, G.K.: The jpeg still picture compression standard. IEEE transactions
  on consumer electronics  \textbf{38}(1),  xviii--xxxiv (1992)

\bibitem{wang2004image}
Wang, Z., Bovik, A.C., Sheikh, H.R., Simoncelli, E.P.: Image quality
  assessment: from error visibility to structural similarity. IEEE transactions
  on image processing  \textbf{13}(4),  600--612 (2004)

\bibitem{wiegand2003overview}
Wiegand, T., Sullivan, G.J., Bjontegaard, G., Luthra, A.: Overview of the h.
  264/avc video coding standard. TCSVT  \textbf{13}(7),  560--576 (2003)

\bibitem{wu2021learned}
Wu, Y., Li, X., Zhang, Z., Jin, X., Chen, Z.: Learned block-based hybrid image
  compression. IEEE Transactions on Circuits and Systems for Video Technology
  (2021)

\bibitem{wu2019detectron2}
Wu, Y., Kirillov, A., Massa, F., Lo, W.Y., Girshick, R.: Detectron2.
  \url{https://github.com/facebookresearch/detectron2} (2019)

\bibitem{wu2018unsupervised}
Wu, Z., Xiong, Y., Yu, S.X., Lin, D.: Unsupervised feature learning via
  non-parametric instance discrimination. In: CVPR. pp. 3733--3742 (2018)

\bibitem{xia2020emerging}
Xia, S., Liang, K., Yang, W., Duan, L.Y., Liu, J.: An emerging coding paradigm
  vcm: A scalable coding approach beyond feature and signal. In: 2020 IEEE
  International Conference on Multimedia and Expo (ICME). pp.~1--6. IEEE (2020)

\bibitem{yang2020improving}
Yang, Y., Bamler, R., Mandt, S.: Improving inference for neural image
  compression. vol.~33, pp. 573--584 (2020)

\bibitem{zhang2016colorful}
Zhang, R., Isola, P., Efros, A.A.: Colorful image colorization. In: ECCV. pp.
  649--666. Springer (2016)

\bibitem{zhao2017pyramid}
Zhao, H., Shi, J., Qi, X., Wang, X., Jia, J.: Pyramid scene parsing network.
  In: CVPR. pp. 2881--2890 (2017)

\end{thebibliography}
% \end{document}

\clearpage
\newpage
\appendix

% ------ Supplementary -------

% \linenumbers
% \pagestyle{headings}
% \mainmatter
% \def\ECCVSubNumber{1954}  % Insert your submission number here

% \title{Supplementary} % Replace with your title

% % INITIAL SUBMISSION 
% % \begin{comment}
% \titlerunning{} 
% \authorrunning{} 
% \author{}
% \institute{}
% % \end{comment}
% % %******************
% \maketitle

\section{Discussion about Image Coding for Machines (ICM)}
% Image coding for machines (ICM) is an emerging field of research, thus here we describe in more detail how it relates to and differs from some fields.
In this section, we describe in detail how the approach we take to tackle the problem of image coding for machines (ICM) differs from those of related tasks.

\subsection{Relationship to Image Coding for Human Perception}

The initial purpose of lossy image compression~\cite{sullivan2012overview,bross2021overview,cheng2020learned} is to ensure the fidelity of the reconstructed image as much as possible. Such fidelity are often measured by objective metrics such as PSNR and MS-SSIM~\cite{hore2010image,wang2004image}. A reconstructed image with a small distortion is supposed to have a good viewing effect.

% To a certain extent, this is subject to the prior knowledge that a reconstructed image with a small distortion are supposed to have a good viewing effect.

% Recently, generative adversarial networks (GAN)~\cite{goodfellow2014generative} are proposed to generate images with high perceptual quality. 

Except the traditional objective metrics, the human eye perception can be well indicated/reflected by the perceptual quality, which is related to the realism of the picture. For example, HiFiC~\cite{mentzer2020high} combines the learning based compression and GAN techniques~\cite{goodfellow2014generative} to get a lossy image compression algorithm with high visual perceptual quality, although the fidelity of compressed image is not very high. 
Moreover, Blau \etal~\cite{blau2018perception} have demonstrated that there exists a trade-off of distortion and perception. Thus, balancing the trade-off of rate, distortion and perception is the goal of lossy compression for humans. 
In contrast, the case of image coding for machines can be regarded as balancing a trade-off of rate and intelligent tasks.
% AI tasks and compression efficiency. 
However, since there exists lots of downstream tasks and even unknown ones, it is difficult to optimize them uniformly. 
Therefore, in this paper, we choose a generalized representation learning method, \ieno, Omni-ICM, to make the learned representation not biased to any task, and general enough for supporting different intelligent tasks.

\subsection{Relationship to Information Bottleneck}
Our solution for ICM that aims to learn the omnipotent feature can also be viewed as a particular instantiation of the more general information bottleneck framework~\cite{tishby2000information,alemi2016deep}. Here we learn the omnipotent representation by maximize the mutual information between our representation and the target of instance discrimination, and meanwhile constrain the mutual information between our representation and the original data. This procedure can be formulated as:

\begin{equation}
\small
    \mathop{\min}_{\boldsymbol{\theta}} I(Z,Y;\boldsymbol{\theta}) \text{ s.t. } I(X,Z;\boldsymbol{\theta}) \leq I_c,
\end{equation}
\noindent where $X$ indicates the original data, $Z$ indicates the latent representation, $Y$ indicates the optimization target and $\boldsymbol{\theta}$ indicates the functions parameterized by $\boldsymbol{\theta}$. And equivalently, with the introduction of a Lagrange multiplier $\beta$ to control the trade-off, it can be formulated as maximize the objective function:

\begin{equation}
    R_{IB}(\boldsymbol{\theta}) = 
    I(Z,Y;\boldsymbol{\theta}) - {\beta}I(Z,X;\boldsymbol{\theta}).
\end{equation}

But differently, our work pays more attentions on how to achieve a good trade-off between compression efficiency and AI tasks generalization, which is not only a naive application or extension of information bottleneck.

\subsection{Relationship to Self-supervised Learning}
\label{sec: relation_SSL}
Methods in self-supervised learning (SSL)~\cite{jing2020self,zhang2016colorful,pathak2016context,noroozi2016unsupervised} are proposed to learn general representations for downstream tasks by solving various pretext tasks on large-scale unlabeled datasets. 
There are mainly two differences between SSL and our method here. One is that the self-supervised learning targets at good initialization weights through pre-training. In subsequent task migration, the entire network is often fine-tuned according to downstream tasks. 
Our method targets at a general representation learning. Once the training is over, the original image is no longer visible to the machines, and replaced by representation extracted from the original image. Thus, the network weights of extracting this representation (backbone head as described in the Section 3.2 of main text) are not allowed to be updated. 
The second point is that SSL has no explicit constrains about entropy but we did, for the reason that we focus on both of the representation ability and the information quantity. In a word, we need to balance the trade-off between generalization and the amount of information of the representation.

\subsection{Relationship to Meta-learning}
Meta-learning~\cite{vanschoren2018meta,hospedales2020meta,snell2017prototypical,finn2017model}, also called as learning-to-learn, provides an alternative paradigm where a machine learning model gains experience over multiple learning episodes - often covering a distribution of related tasks. The experience of mentioned procedure would help improve the future learning performance.
Similar to SSL mentioned in Section \ref{sec: relation_SSL}, the pre-trained model are often fine-tuned for downstream tasks with all parameters updated. In addition, meta-learning also does not explicitly need a representation with low entropy thus easy to compressing.

\section{Architecture Details}

\begin{figure}[t]
  \vspace{-2mm}
  \centerline{\includegraphics[width=1.0\linewidth]{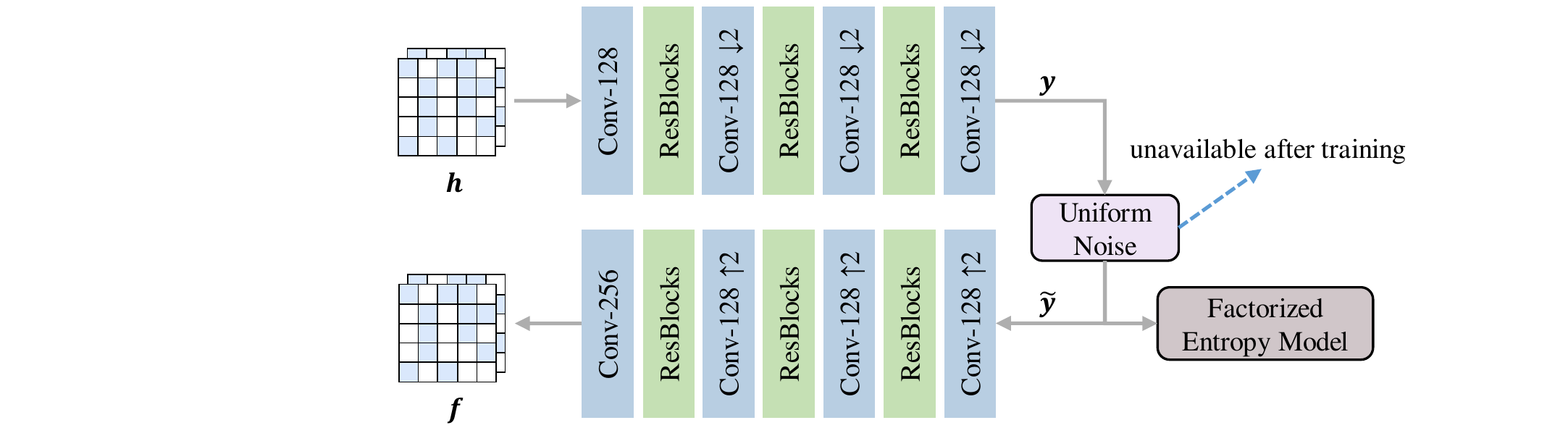}}
    \vspace{-2mm}
    \caption{The architecture of information filtering (IF) module. Each ``Resblocks'' in the figure is stacked by three ResBlocks. And a ResBlock consists of two conlutional layers (with $3\times3$ kernel size, 128 input channels and 128 output channels) which involved by a single shortcut.
     }
\label{fig:arch_IF_module}
\end{figure}

\noindent\textbf{Information Filtering (IF) Module.}
The detailed architecture of the information filtering (IF) module illustrate in Figure \ref{fig:arch_IF_module}. Extra residual blocks are used to increase receptive filed and improve non-linear transformation capability~\cite{cheng2020learned}. As for the size of IF module, compared with baseline (ResNet-50) with 25.56M parameters, our method just adds an additional information filter (IF) module, with only 8.24M parameters increased. 

\begin{figure}[t]
  \centerline{\includegraphics[width=1.0\linewidth]{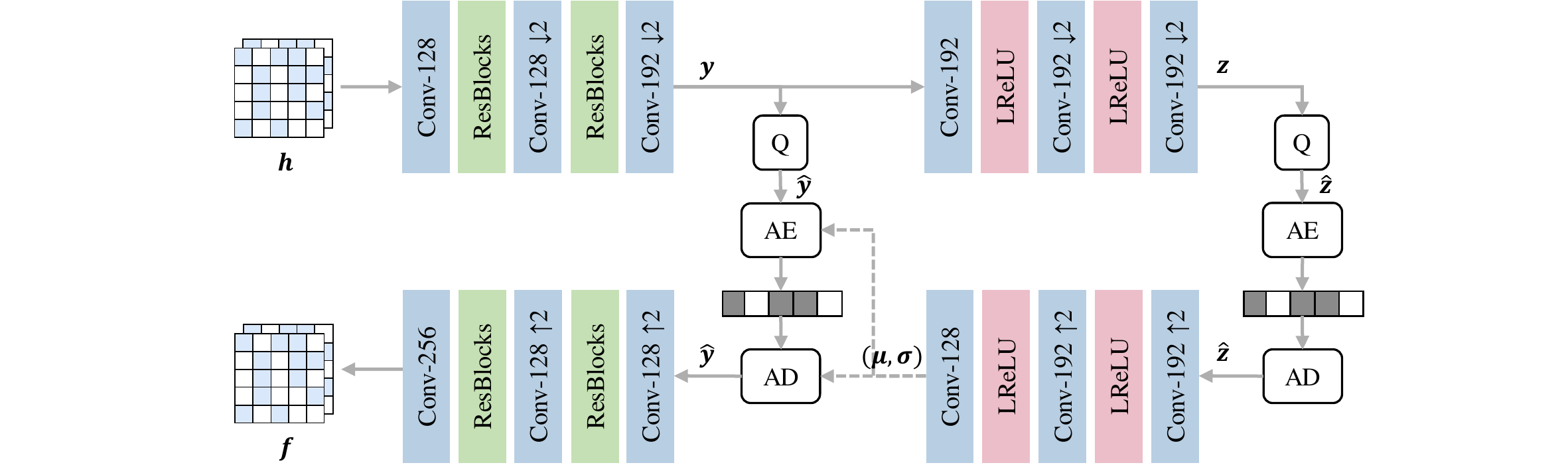}}
  \vspace{-2mm}
    \caption{The architecture of the learning-based feature compressor. Each ``Resblocks'' in the figure is stacked by three ResBlocks. And a ResBlock consists of two conlutional layers (with $3\times3$ kernel size, 128 input channels and 128 output channels) which involved by a single shortcut. LReLU indicates LeakyReLU.
     }
\label{fig:arch_compressor}
\vspace{-2mm}
\end{figure}

\noindent\textbf{Feature Compression Codec.}
The detailed architecture of the learning-based feature compression codec is illustrated in Figure \ref{fig:arch_compressor}. Note that, here the several residual blocks are used to increase receptive filed and improve the entire rate-distortion performance. 

\section{More Experimental Results}
\begin{figure*}
\vspace{-2mm}
  \centerline{\includegraphics[width=1.0\linewidth]{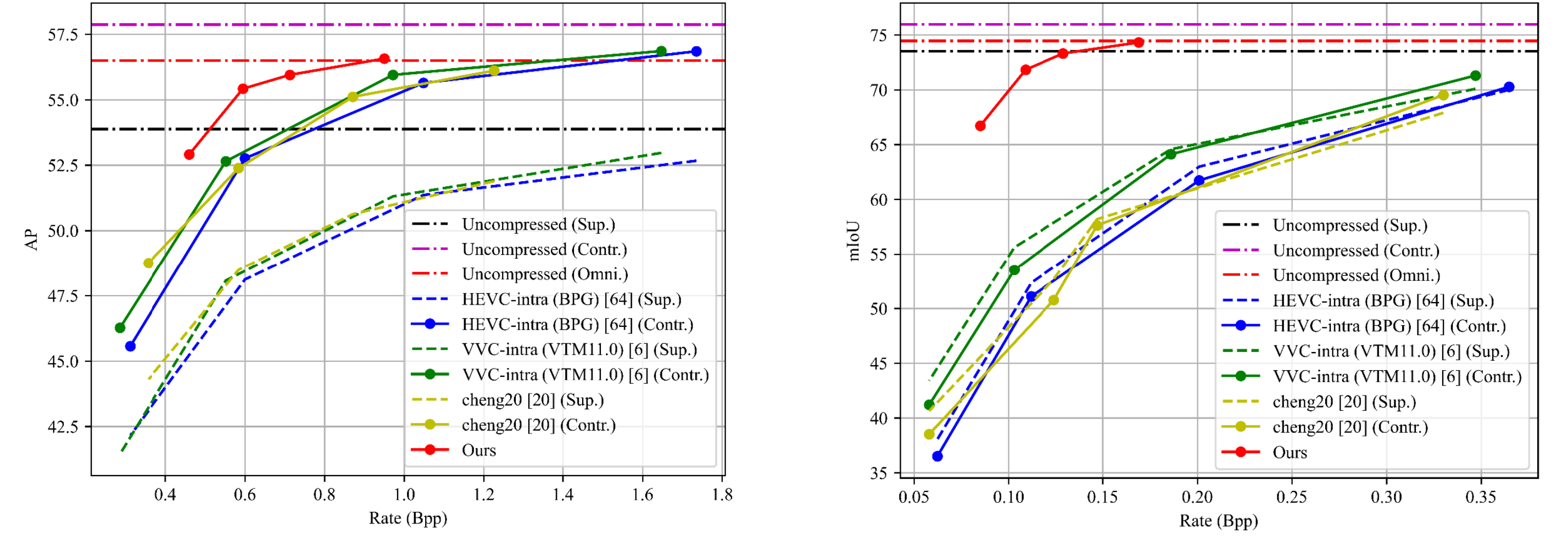}}
  \vspace{-2mm}
    \caption{
    \textbf{
    Object detection on PASCAL VOC (left) and semantic segmentation on Cityscapes (right). }
    Dotted lines indicate the results of uncompressed data as input. 
    Dashed lines indicate the results of fine-tuning with ImageNet supervised pre-training weights.
     }
\label{fig:full_det_voc_cityscapes_segmentation}
\vspace{-2mm}
\end{figure*}

\begin{figure*}
\vspace{-2mm}
  \centerline{\includegraphics[width=1.0\linewidth]{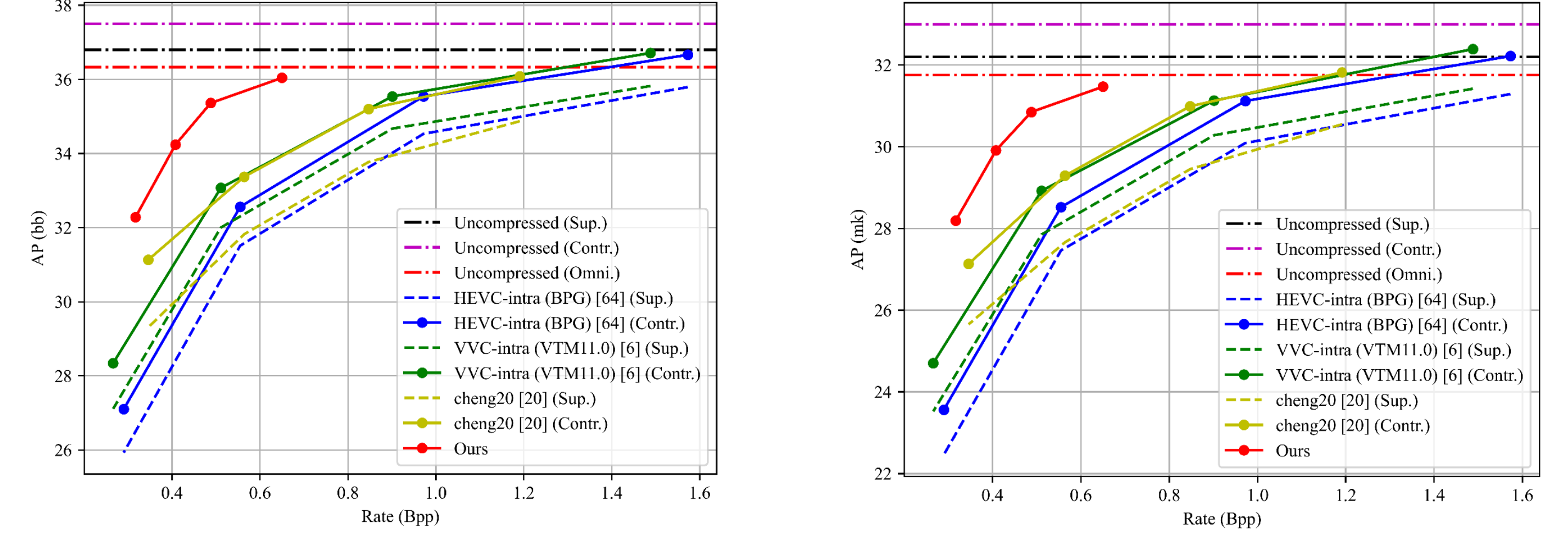}}
  \vspace{-2mm}
    \caption{
    \textbf{Object detection and instance segmentation on MS coco.}
    Dotted lines indicate the results of uncompressed data as input. 
    Dashed lines indicate the results of fine-tuning with ImageNet supervised pre-training weights.
     }
\label{fig:full_coco_instance}
\vspace{-2mm}
\end{figure*}

\begin{figure*}
\vspace{-2mm}
  \centerline{\includegraphics[width=1.0\linewidth]{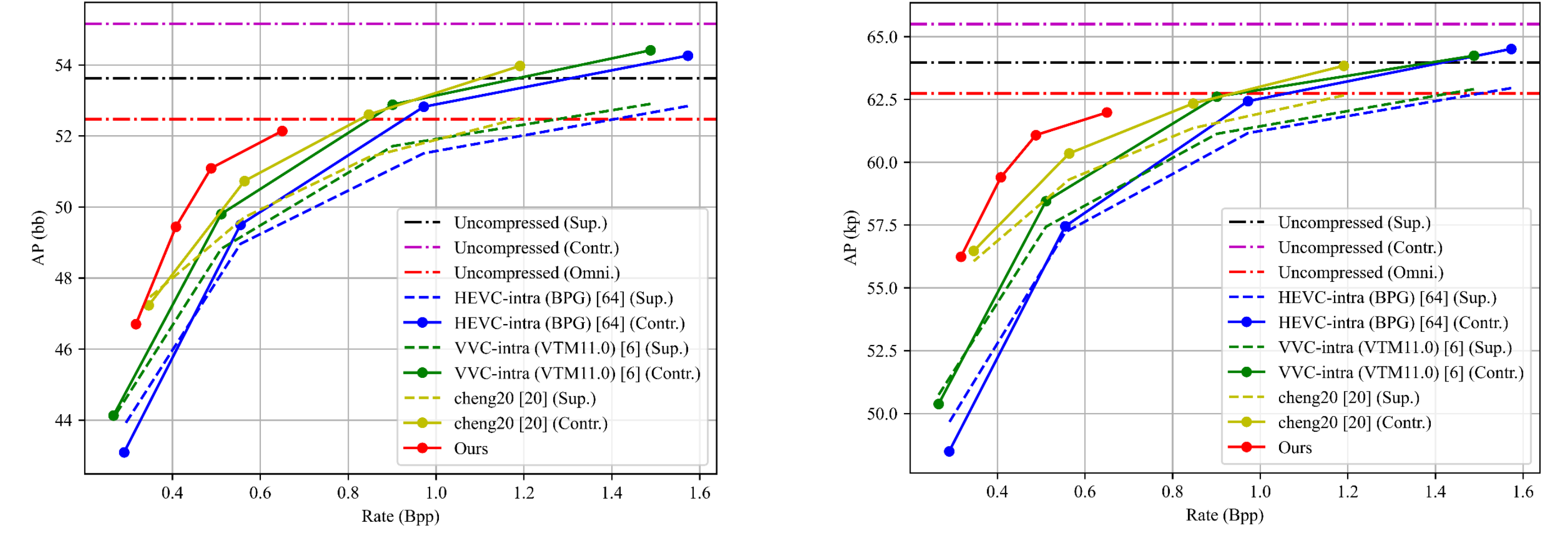}}
  \vspace{-2mm}
    \caption{
    \textbf{Pose estimation on MS COCO.}
    Dotted lines indicate the results of uncompressed data as input. 
    Dashed lines indicate the results of fine-tuning with ImageNet supervised pre-training weights.
     }
\label{fig:full_coco_keypoint}
\vspace{-2mm}
\end{figure*}

\begin{figure*}
\vspace{-2mm}
  \centerline{\includegraphics[width=1.0\linewidth]{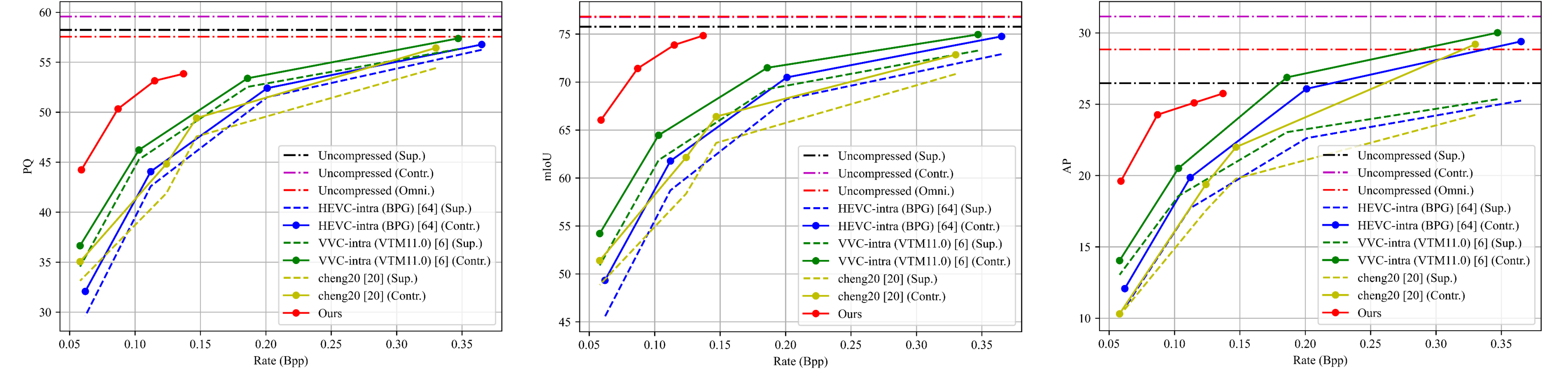}}
  \vspace{-2mm}
    \caption{
    \textbf{Panoptic segmentation on Cityscapes.}
    Dotted lines indicate the results of uncompressed data as input. 
    Dashed lines indicate the results of fine-tuning with ImageNet supervised pre-training weights.
     }
     \vspace{-2mm}
\label{fig:full_panoptic_seg}
\end{figure*}

\vspace{-1mm}
More experimental results are illustrated in Fig.~\ref{fig:full_det_voc_cityscapes_segmentation}, \ref{fig:full_coco_instance}, \ref{fig:full_coco_keypoint}, \ref{fig:full_panoptic_seg}. 
% We additionally compare our method with two kinds of competitors.
% They are baselines (noted as ``Uncompressed'' in figures) and supervised fine-tuning.
% The former means that uncompressed images or features are input to task models for evaluation.
% And the latter means that we additionally train task models with the ImageNet pre-trained weights as initialization and evaluate on them.
We additionally compare our method with the competitors of supervised fine-tuning. For this case, we train task models with the ImageNet pre-trained weights as initialization and evaluate on them.
As shown in these figures, the performances of baselines that fine-tuning on contrastive learning pre-trained models are better than those of fine-tuning on supervised learning pre-trained models. 
And the baselines of our Omni-ICM have a drop compared with fully contrastive pre-training. The degrees of decline vary according to the datasets and tasks.
As for the case of coding for intelligent tasks (the curve part of the paradigms), results on task models fine-tuning on contrastive pre-training are better than those fine-tuning on supervised pre-training, and our methods performs better than both of them.

%% results with baselins and supervised pre-trained

% \section{Training details of reconstruction from features}

\section{Details about Feature Reconstruction}

\begin{figure}
    \vspace{-2mm}
  \centerline{\includegraphics[width=1.0\linewidth]{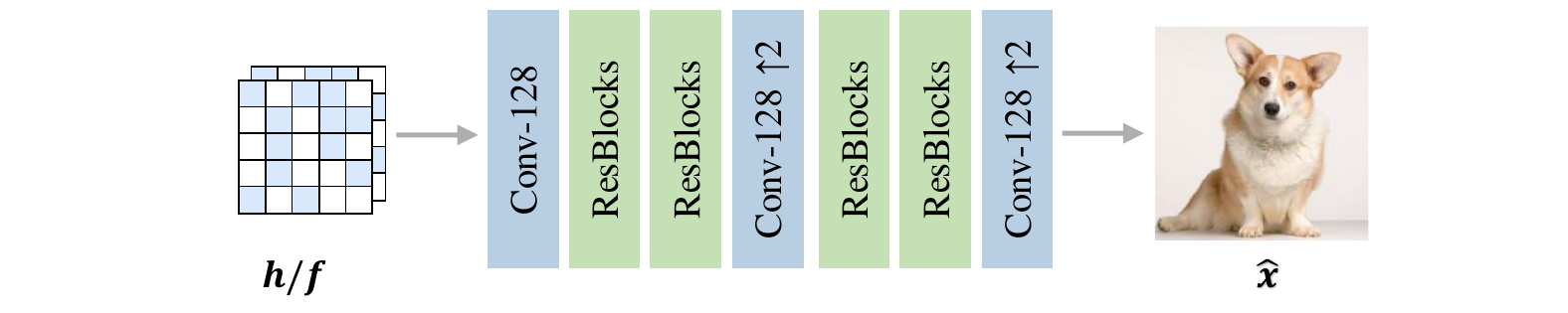}}
  
    \caption{The architecture of decoder for reconstruction from features. Each ``Resblocks'' in the figure is stacked by three ResBlocks. And a ResBlock consists of two conlutional layers (with $3\times3$ kernel size, 128 input channels and 128 output channels) which involved by a single shortcut.
     }
\label{fig:arch_reconstruction_decoder}
\vspace{-2mm}
\end{figure}

\noindent\textbf{Decoder Architecture.} The architecture of the decoders (mentioned in Section 4.6 of the text) for reconstruction from features are stacked by convolutional layers and ResBlocks, which is illustrated in Fig. \ref{fig:arch_reconstruction_decoder}.
Residual blocks are also used to increase receptive filed and improve non-linear transformation capability. 
These two decoders share the same architecture and training schedule.
We train them for $200,000$ iterations with batch size of 16. Adam optimizer\cite{kingma2014adam} is employed and the learning rate is set as $5\times10^{-5}$. Data augmentation is $256\times256$ random cropping.

\begin{figure*}
\vspace{-4mm}
  \centerline{\includegraphics[width=1.0\linewidth]{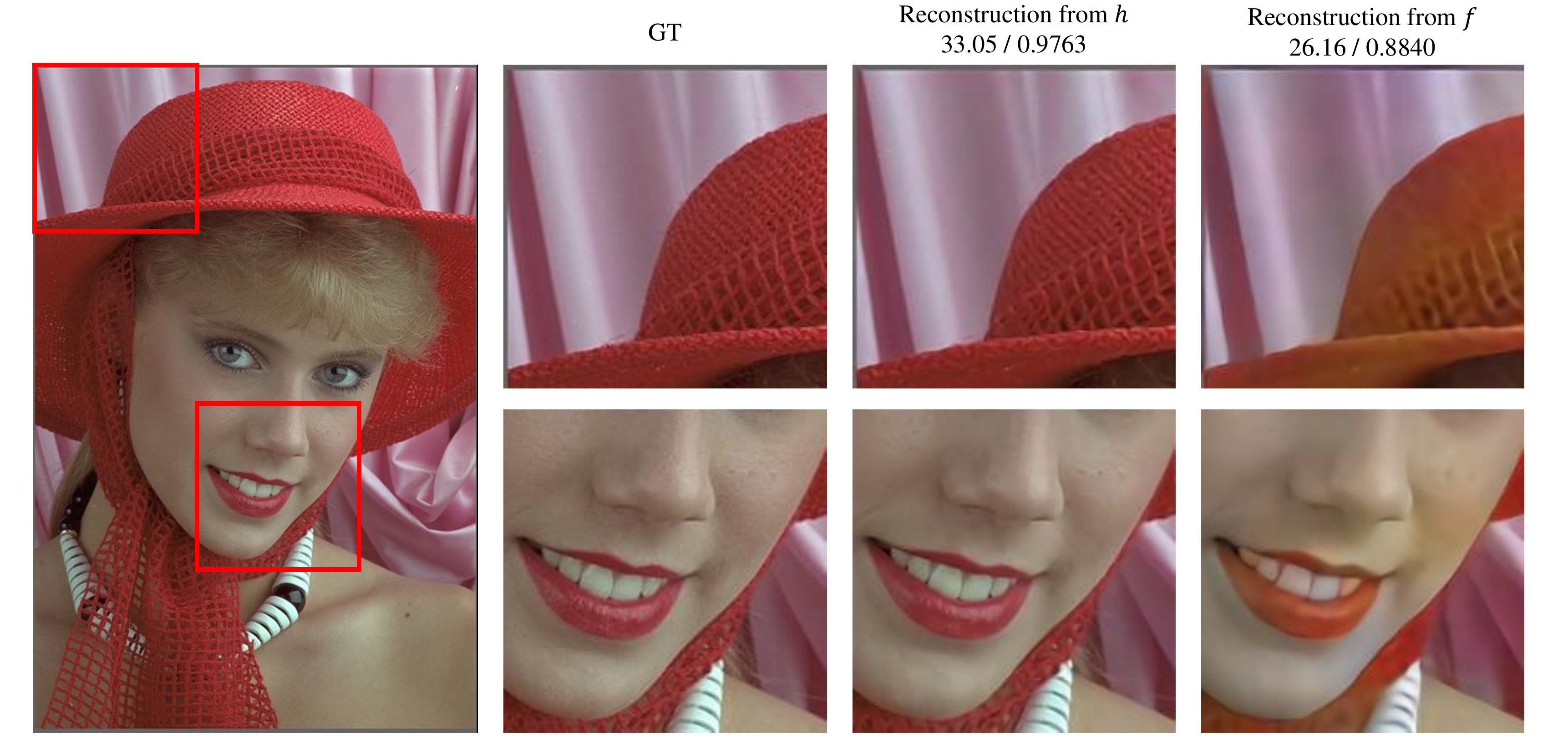}}
  \vspace{-2mm}
    \caption{Reconstruction of features before and after IF module on Kodak 4 image. The numbers on the top of the crop images indicate PSNR (dB) / MS-SSIM of an entire image.
     }
\label{fig:qualitative_kodak04}
\vspace{-2mm}
\end{figure*}

\begin{figure*} 
\vspace{-4mm}
  \centerline{\includegraphics[width=1.0\linewidth]{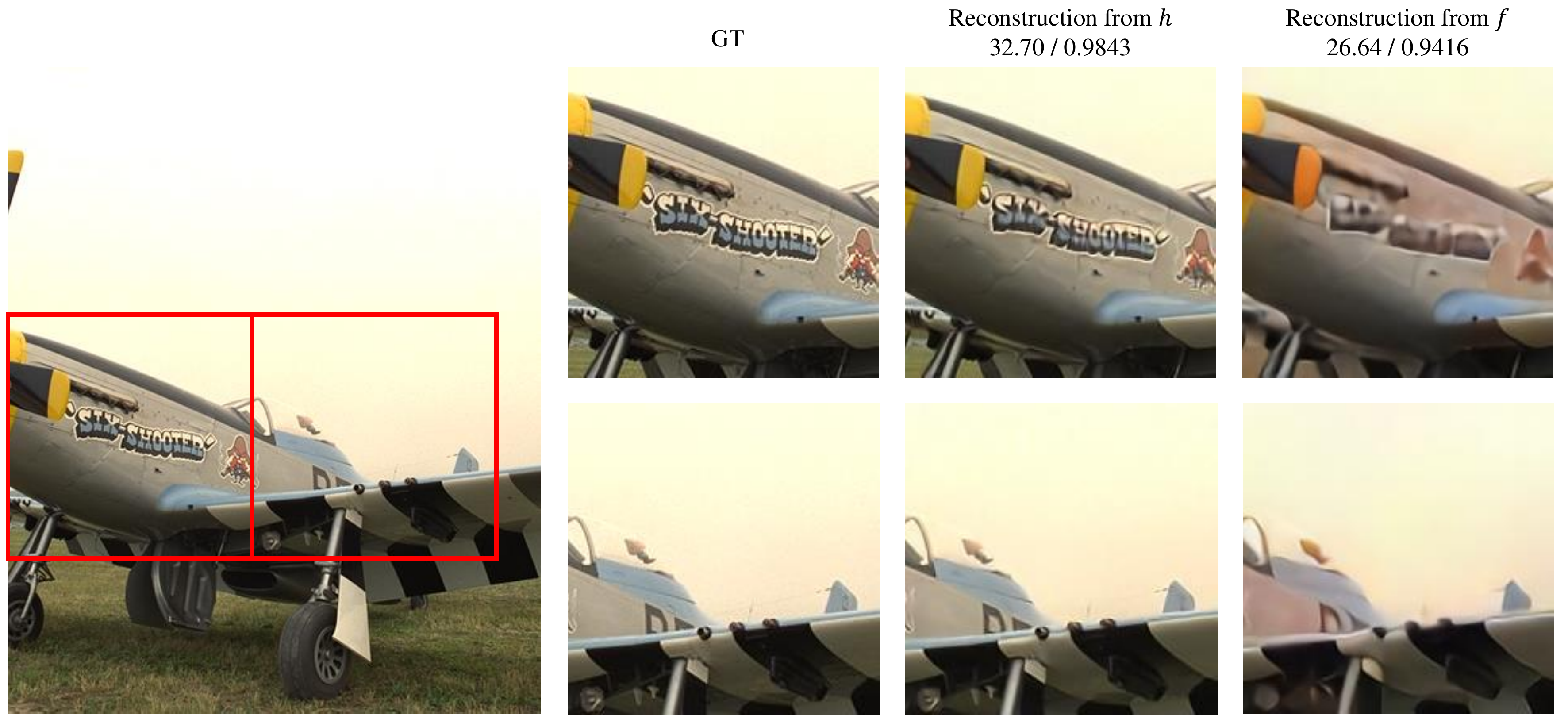}}
  \vspace{-2mm}
    \caption{
    Reconstruction of features before and after IF module on Kodak 20 image. The numbers on the top of the crop images indicate PSNR (dB) / MS-SSIM of an entire image.
     }
\label{fig:qualitative_kodak20}
\vspace{-2mm}
\end{figure*}

\begin{figure*} 
  \centerline{\includegraphics[width=1.0\linewidth]{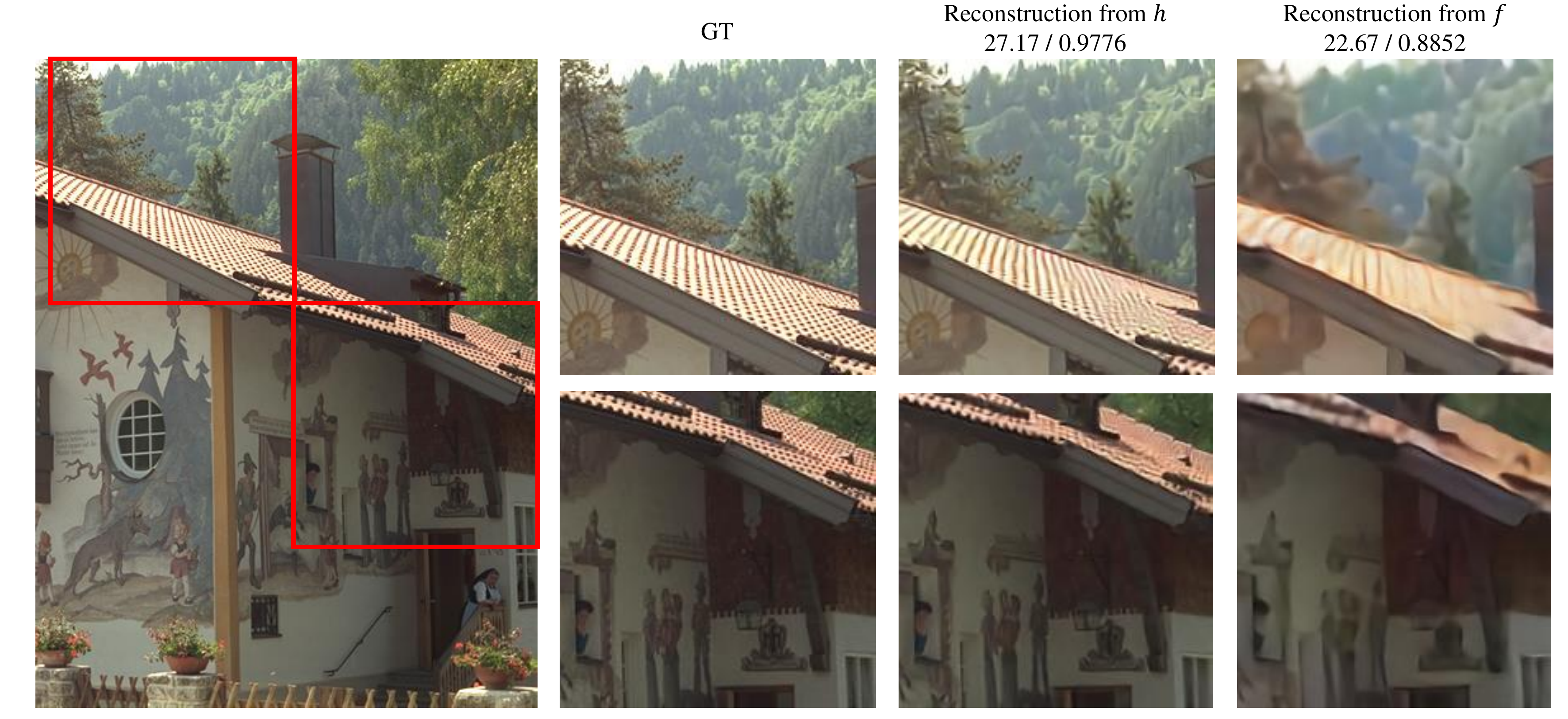}}
  \vspace{-2mm}
    \caption{
        Reconstruction of features before and after IF module on Kodak 24 image. The numbers on the top of the crop images indicate PSNR (dB) / MS-SSIM of an entire image.
     }
\label{fig:qualitative_kodak24}
\vspace{-2mm}
\end{figure*}

\noindent\textbf{More reconstruction results.}
Fig. \ref{fig:qualitative_kodak04}, \ref{fig:qualitative_kodak20}, \ref{fig:qualitative_kodak24} show more results of reconstruction of features before and after IF module on several Kodak images. We can see that the images reconstructed from $h$ contain slight color different, and textures are relatively complete. But, the images that reconstructed from $f$ suffer from more obvious color jitter and texture distortion. This indicate that our information filtering (IF) module indeed filter out these color and texture information that have a slight influence on intelligent analytics.

\vfill

\end{document}